\documentclass[review]{elsarticle}
\usepackage[T1]{fontenc}
\usepackage{lmodern}
\usepackage{lineno,hyperref}
\modulolinenumbers[5]

\usepackage[utf8]{inputenc} % allow utf-8 input
\usepackage[T1]{fontenc}    % use 8-bit T1 fonts
\usepackage{hyperref}       % hyperlinks
\usepackage{url}            % simple URL typesetting
\usepackage{booktabs}       % professional-quality tables
\usepackage{amsfonts}       % blackboard math symbols
\usepackage{nicefrac}       % compact symbols for 1/2, etc.
\usepackage{microtype}      % microtypography
\usepackage{lipsum}
\usepackage{xcolor}
\usepackage{amssymb}

\usepackage[a4paper, text={16cm,21cm},top=2.5cm, left=2.5cm, right=2.5cm, includeheadfoot, headheight=1cm]{geometry}

\usepackage{amsmath}

\usepackage{algorithm}
\usepackage[noend]{algpseudocode}
\usepackage{adjustbox}
\usepackage{booktabs} % To thicken table lines

\usepackage{amssymb}
\usepackage{multirow}
\usepackage{lscape}

\journal{Journal of \LaTeX\ Templates}

\begin{document}
\begin{frontmatter}
\title{Random Machines: A bagged-weighted support vector model with free kernel choice}

\author{Anderson Ara}
\address{Department of Statistics\\
       Federal University of Bahia\\
       Salvador, Bahia, Brazil}
       
\author{Mateus Maia}
\address{Department of Statistics\\
       Federal University of Bahia\\
       Salvador, Bahia, Brazil}
       
\author{Samuel Mac\^edo}
\address{Department of Natural Sciences and Mathemathics\\
       Federal Institute of Pernambuco\\
       Recife, Pernambuco, Brazil}
                     
\author{ Francisco Louzada}
\address{Institute of Mathematical and Computer Sciences \\
       University of S\~ao Paulo \\
       S\~ao Carlos, S\~ao Paulo, Brazil}

\begin{abstract}
Improvement of statistical learning models in order to increase efficiency in solving classification or regression problems is still a goal pursued by the scientific community. In this way, the support vector machine model is one of the most successful and powerful algorithms for those tasks. However, its performance depends directly from the choice of the kernel function and their hyperparameters. The traditional choice of them, actually, can be computationally expensive to do the kernel choice and the tuning processes. In this article, it is proposed a novel framework to deal with the kernel function selection called Random Machines. The results improved accuracy and reduced computational time. The data study was performed in simulated data and over 27 real benchmarking datasets.
\end{abstract}


\end{frontmatter}

\section{Introduction}
The application and development of statistical learning methods is currently an important and significant research field in science. The supervised machine learning techniques have numerous applications in classification tasks ranging from cancer diagnostics and prediction \cite{sato2019machine}, speech recognition \cite{mokgonyane2019automatic}, text classification \cite{burdisso2019text,kim2005dimension} and financial fraud detection \cite{dighe2018detection}. The variety of methods has been used in the field is huge, but one of them emerges, the Support Vector Machine (SVM). The SVM \cite{cortes1995support} is the youngest well established and successful in traditional learning methods. Smola \cite{smola2000advances} presented some good proprieties of this learning algorithm, including good generalization capacity, high efficiency in prediction tasks, beyond the convexity of the objective function which guarantees a global minimum. Some works present the superiority of the SVM when compared with other supervised learning benchmarking techniques, highlighting good accuracy results \cite{cueto2019comparative,thanh2018comparison,shah2018performance}.

At the same time, the ensemble methods have been gaining more strength as a tool to improve the accuracy in classification models. The combination of singular models can enhance predictive power and increase its generalization capacity \cite{van2007improved}. There are two main classes of ensemble algorithms: bagging \cite{breiman1996bagging} that uses independent bootstrap samples to create multiple models and built a final classifier combining them, and boosting algorithms \cite{freund1999short} that built sequential models in order to assign different weights relying on their performance.

The literature already proposed bagging methods jointly with the support vector machine algorithm \cite{kim2002support} as a methodology of increasing its accuracy.  Wang et al. \cite{wang2009empirical} realized an empirical study of Bagged SVM and showed that the technique performs as well or better than other methods with a relatively higher generality. Moreover, different applications of bagged SVM are reported, e.g breast cancer prediction \cite{huang2017svm,wang2018support}, credit score modelling \cite{zhou2010least}, gene detection \cite{tong2013ensemble}, spatial prediction of landslides \cite{pham2018bagging}, bacterial transcription start sites prediction \cite{gordon2005improved} , text speech recognition \cite{lei2006ensemble} and membership authentication \cite{pang2003membership}.

Despite the diverse number of works that present the bagging based on support vector machine classifiers, none of them presents an optimal framework to choose which kernel function will be used in the ensemble classifier. The choice of the kernel function, as their hyperparameters, has a crucial impact on the accuracy of the technique \cite{jebara2004multi}. Generally, this selection is supported by a grid-search that runs all functions and their parameters combinations in order to select which one has the lowest generalization error rate.  Random Search \cite{bergstra2012random} is another approach to tuning the hyperparameters, where the parameters configurations are randomly chosen until a particular budget B is exhausted. Beside these, Tree-Structured Parzen Estimator \citep{bergstra2011algorithms}, and Simulated Annealing \citep{kirkpatrick1983optimization} are optimization structures used in tuning workflow too. However, all of them can be computationally expensive and slow, making it infeasible to use.

The kernel methods, \textit{e.g:} Kernelized Support Vector Machines, could be considered as non parametric machine learning models which are useful to capture the non-linear behaviour, beyond their strong theoretical proprieties. However, they have some problems to be applied to large scale datasets since their time and memory demand, that is at least $n^{2}$, where $n$ is the the number of observations. Currently works, solve the problem of computational limitations through the use of Nystr\"om method \cite{williams2001using,smola2000sparse} or random features. Both of them have their specific versions for support vector machine \cite{sun2018but, li2016fast}, and represents a solid advance in those techniques.

This work introduces a novel method that presents a solution for the choice of kernel function to be used in the bagged supported vector machine, in order to give an alternative to the open problem of hyperparameters' selection, with adequate computational time and robust accuracy power, hereafter, the Random Machines (RM). The method received this name because it uses random kernel choice for each model that composes the bagged support vector machine method, besides proposing weights to these classifiers, increasing the accuracy and lowering the correlation of the final model. The result was validated over simulation studies, and on 27 different benchmarking datasets.

The following paper is organized on the ensuing outline. The Section 2 presents a theoretical description about the support vector machine method, proposed by \cite{cortes1995support},the challenges on the selection of hyperparameters and some traditional kernel functions; Section 3 presents a general description of the bagging algorithm and bagged SVM; Section 4 presents how the proposed Random Machines (RM) approach works in detail, followed by the simulations studies in Section 5, as well as the applications in real data in Section 6. Section 7 shows an empirical justification of how the method works that proves the consistency of the technique. Finally, in Section 8, final considerations, regarding the improvements and limitations that could be explored in this novel approach.
\label{sec1}

\section{Support Vector Machine}
The support vector machines \cite{cortes1995support}, have been introduced for solving classification problems. The overall idea of the technique is to calculate a hyperplane which separates observations between two classes, maximizing the distance between the support vectors. 

Supposing a database given by \{$\mathbf{x_{i}},\mathbf{y_{i}}$\},  $y_{i}=\{-1,1\}$, i=1,$\dots$, $n$, where $n$ is the number of observations. The $y_{i}=1$ represents that the observation belongs to a positive class, while $y_{i}=-1$ the negative one. Therefore, the hyperplane that accurately separate these two classes is given by
\begin{equation}
    \centering
    \mathbf{w} \boldsymbol{\cdot} \mathbf{x + b}=0
\end{equation}

In order to find such hyperplane  the estimation of $\mathbf{w}$ and $\mathbf{b}$ is made in order to maximize the distance between the support vectors \cite{boser1992training,cortes1995support}, following the restrictions of $ y_{i} (w \boldsymbol{\cdot} \mathbf{x_{i}} + b) \ge 1$, if $y_{i}$ belongs to the positive class $y_{i}=1$, and $ y_{i} (w \boldsymbol{\cdot} \mathbf{x_{i}} + b)\ge -1$, otherwise $y_{i}=-1$. These equations are expressed by
\begin{equation}
    \centering
    y_{i} (w \boldsymbol{\cdot} \mathbf{x_{i}} + b) -1 \ge 0
        \label{eq:rest}
\end{equation}

The distance is given by $\frac{2}{||w||}$, to maximize it is necessary to solve a convex problem given by 
\begin{equation}
    \centering
    \min \frac{1}{2}||w||^{2}
    \label{eq:min_func}
\end{equation}
following the constraints given by the Equation (\ref{eq:rest}). The cost function which will be minimized is defined by the Lagrangian Multipliers, in Equation (\ref{eq:work_func}).
\begin{equation}
    L(\mathbf{w,b,}\boldsymbol{\alpha})=\frac{1}{2}||w||^{2}-\sum_{i=1}^{n}\alpha_{i}[y_{i} (w \boldsymbol{\cdot} \mathbf{x_{i}} + b) -1 ]
    \label{eq:work_func}
\end{equation}
where $\alpha_{i}$ is the Lagrangian Multiplier.

There are cases where the training data cannot be separated without error, as pointed out by  \cite{cortes1995support}. In such a case, it is needed to construct a soft margin separator by inputting slack variables ($\varepsilon_{i}$). Therefore, a transformation in the Equation (\ref{eq:min_func}) was needed \cite{cortes1995support}, and then, it becomes 
\begin{equation}
    \centering
    \min \frac{1}{2}||w||^{2} + C\sum_{i=1}^{n}\varepsilon_{i}
\end{equation}
where $C\ge0$ is a regularization parameter. The constraints become  $y_{i} (w \boldsymbol{\cdot} \mathbf{x_{i}} + b)-(1 -\varepsilon_{i}) \ge 0$ and $\varepsilon_{i} \ge 0 $ for $i=1,\dots,n$. And the cost function, which will be minimized, becomes
\begin{equation}
    L(\mathbf{w},\mathbf{b},\boldsymbol{\alpha},\mathbf{r})=\frac{1}{2}||w||^{2} + C\sum_{i=1}^{n}\varepsilon_{i}-\sum_{i=1}^{n}\alpha_{i}[y_{i} (w \boldsymbol{\cdot} \mathbf{x_{i}} + b) -1 +\varepsilon_{i}] -\sum_{i=1}^{n}r_{i}\varepsilon_{i}.
\end{equation}

The solution, considering the Lagrangian Dual Optimization for the soft margin problem \cite{fletcher1987practical},  is given by
\begin{equation}
    \centering
    \max_{\boldsymbol{\alpha}} \left(\sum_{i}^{n}\alpha_{i}-\frac{1}{2}\sum_{i}^{n}\sum_{j}^{n}\alpha_{i}\alpha_{j}y_{i}y_{j} \mathbf{x_{i}} \boldsymbol{\cdot} \mathbf{x_{j}} \right)
    \label{eq:dual}
\end{equation}

\begin{equation*}
    \text{s.t}=  
\begin{cases}
  \sum_{i}^{n}\alpha_{i}y_{i}=0, \\
 C\ge\alpha_{i}\ge0,
\end{cases}
\end{equation*}
with $i=1,\dots,n$.

This approach of SVM works well to linearly classification groups and problems. In the presence of non-linearity, it may be used trick kernels, based in Mercer's Theorem. Instead of considering the input space, it's considered higher feature spaces, where the observations could be linearly separable through the following function $K\mathbf{(x_{i},x_{j})}=\phi(\mathbf{x_{i}})\cdot\phi(\mathbf{x_{j}})$  that replaces the inner product in Equation (\ref{eq:dual}).

The functions $K(x,y)=\phi(\mathbf{x)}\cdot\phi(\mathbf{y})$ are defined as the semidefinite kernel functions \cite{courant1953methods}. Several types of kernel functions are employed in different classification tasks. The choice of distinct kernels functions provide different nonlinear mappings, and the performance of the resulting SVM often depends on the appropriate choice of the kernel \cite{jebara2004multi}. Some works that compare the efficiency for each kernel function, which is used for each classification model \cite{hussain2011comparison,min2005bankruptcy}, demonstrating that select the kernel function is an important aspect of obtaining the best model.  There are kernel functions in the general framework for SVM, which were used in this paper, that are considered the most common. They are presented in Table \ref{tab:kernel}.

\begin{table}[H]
\centering
\caption{Kernel Functions.}
\begin{tabular}{|l|c|c|}
\hline
\textbf{Kernel}   & \textbf{K(x,y)} & \textbf{Parameters}\\  \hline
Linear Kernel     & $\gamma(x\cdot y)$  & $\gamma$              \\
Polynomial Kernel & $(\gamma( x\cdot y))^{d}$ & $\gamma,d$   \\
Gaussian Kernel   & $e^{-\gamma||x-y||^2}$ & $\gamma$ \\
Laplacian Kernel  & $e^{-\gamma||x-y||}$ & $\gamma$ \\ \hline
\end{tabular}
\label{tab:kernel}
\end{table}
 in which $\gamma \in \mathbb{R^{+}}, d \in \mathbb{N}$.
 
 \vspace{0.25cm}
 
Nevertheless, find out which is the best kernel by grid search, or other exhaustive methods, can be an expensive and appalling computational problem \cite{chapelle2000model}. In order to deal with this issue, many works have tried to develop a methodology which can improve the selection of the best kernel function \cite{jebara2004multi,ayat2005automatic,wu2009novel,friedrichs2005evolutionary,cherkassky2004practical}. In this work we propose a novel approach which makes unnecessary to perform a grid search, or other tuning algorithm, to choose a single specific kernel function when using the trick kernel.

\section{Bagging}

Bagging is an abbreviation of Bootstrapping Aggregation, which was firstly proposed by Breiman \cite{breiman1996bagging}. Bagging is an ensemble method that can be used for different prediction tasks. In general, the Bootstrapping Aggregating generates datasets by random sampling with replacement from the training set with the same size $n$, also known as bootstrap samples. Then, each model $h_{j}(x_{i})$ is trained independently for each bootstrapping sample $j$, $\forall j \in\{1,\dots,B\}$. The final bagging model, for binary classification tasks, is given by the following equation,
\begin{equation}
    H(\mathbf{x})=sign\left(\sum_{i=1}^{B}h_{i}(\mathbf{x})\right),
    \label{eq:bagging}
\end{equation}
where $h_{i}(\mathbf{x})$ is the model generated to each bootstrap sample from $i=1,,\dots,B$, and $B$ is the number of bootstrap samples. 

Another critical feature of Bagging classifier is the out of bag samples \cite{breiman1996bagging}. For each bootstrap sample, almost $\frac{1}{3}$ of observations are not included. Therefore, those observations can be used as a test sample since they were not used to train the bootstrap models.

%It's  know that combining multiple classifiers, e.g ensemble methods, often increase the efficiency when compared to a single classifier \cite{duin1998combining}. 

\subsection{Bagging SVM}

In the bagging classifier, the function $h_{i}(\mathbf{x})$ from (\ref{eq:bagging}) can be any model. One possibility is to use the SVM as the base classifier \cite{kim2002support} in order to improve it is accuracy. As we already have seen, the applications of the bagged SVM for predictive tasks are wide, and empirical studies \cite{wang2009empirical} demonstrated that the bagged version of the support vector machine algorithm increased the accuracy and it is generalization capacity. Moreover, some of them already presented some modifications using the SVM in bagging context as \cite{lin2008support}, and others implemented some libraries as \textit{EnsembleSVM}, that make it shorten to use simple ensemble methods with SVM \cite{claesen2014ensemblesvm}.

{
Despite the numerous works using bagged SVM, none of them present a general framework to deal with the choice of the best kernel function, choosing it by trial evaluation or by a grid search. As this proceeding is computationally expensive \cite{chapelle2000model}, this paper proposed a novel bagging approach that can overcome the difficult to choose the best kernel function, besides showing an improvement in the accuracy of classification models by combining several different SVM models by varying the kernel functions: the Random Machines, exposed in next section.
}

\section{Random Machines}
Given a training set $\{(x_{i},y_{i})\}_{i=1}^{N}$ with $\mathbf{x_{i}} \in \mathbb{R}^{p}$ and $y_{i} \in \{-1,1\}$, $\forall i=1,\dots,n$. The kernel bagging method initialize by training single models $h_{r}(\mathbf{x})$, where $r=1,\dots,R$, where $R$ is the  total number of different kernel functions that could be used in support vector machine models. For example, if $R=4$ a possible choice is define $h_{1}$ as SVM with \textit{Linear kernel} , $h_{2}$ as SVM with \textit{Polinomial kernel}, $h_{3}$ as SVM with \textit{Gaussian kernel} and $h_{4}$ as SVM with \textit{Laplacian kernel}. 

Each model is validated for the test set $\{(x_{k},y_{k})\}_{k=1}^{L}$, and the accuracy $ACC_{r}$ is calculated for each model, $\forall r=1,\dots,R$, in which $R$ means the numbers of kernel functions that will be used. Therefore, the probabilities, $\lambda_{r}$, is given by the Equation (\ref{eq:prob}) for each kernel function

\begin{equation}
    \lambda_{r}=\frac{ln\left(\frac{ACC_{r}}{1-ACC_{r}}\right)}{\sum_{i=1}^{R}ln\left(\frac{ACC_{i}}{1-ACC_{i}}\right)}, 
    \label{eq:prob}
\end{equation}
with $\forall r=1,\dots,R$.

Afterwards, is sampled $B$ bootstrap samples from the training set. A support vector machine model $g_{k}$ is trained for each bootstrap sample, $k=i,\dots,B $ and the kernel function that will be used for $g_{k}$ will be determined by a random choice with probability $\lambda_{r}, \forall r=1,\dots,R$.
The probabilities $\lambda_{r}$ are higher if determined kernel function used in $h_{r}(\mathbf{x})$ predicted correctly observations from test set. Consequently, the kernel functions with higher accuracy will appear often when the random kernel selection for each bootstrap model is made. If any kernel function applied in $h_{r}(\mathbf{x})$ does not do better than a random choice, then $ACC_{r}$ is closer to $0.5$ and the probability of select that kernel function is next to zero.

Subsequently, a weight $w_{i}$ is assigned to each bootstrap model calculated for  $g_{i}$ $\forall i=1,\dots,B$. The weight is given by the Equation (\ref{eq:weight}).

\begin{equation}
    w_{i}=\frac{1}{(1-\Omega_{i})^2}, \hspace{0.3cm} i=1,\dots,B ,
    \label{eq:weight}
\end{equation}

where $\Omega_{i}$ is the accuracy of model's prediction $g_{i}$ calculated on Out of Bag Sample ($OOBG_{i}$) obtained from $i$ bootstrap sample $\forall i=1,\dots,B$ as test sample.

The final classification is held in Equation (\ref{eq:final_class}).

\begin{equation}
    G(\mathbf{x_{i}})=sign \left( \sum_{j}^{B}w_{j}g_{j}(\mathbf{x_{i}})\right), \hspace{0.3cm} i=1,\dots,N.
    \label{eq:final_class}
\end{equation}
 All the modeling process is summed up in the pseudo-code exposed in Algorithm \ref{alg:RM}.
\begin{algorithm}[H]
\caption{Random Machines}
\begin{algorithmic} 
    \State{\bf{Input}: Training Data, Test Data, B, Kernel Functions}
    \For{each $Kernel Function_{r}$}
        \State Calculate the model  $h_{r}$
        \State Calculate the accuracy $\alpha_{r}$
    \EndFor
    \State \textbf{Calculate} the probabilities $\lambda_{r}$
    \State \textbf{Generate} B bootstrap samples
    \For{b in B}
        \State Model $g_{b}( \mathbf{x_{i}})$ by sampling a kernel function with probability
        $\lambda_{r}$
        \State Assign a weight $\Omega_{b}$ using $OOBG_{b}$ samples.
    \EndFor
    \State \textbf{Calculate} $G(\mathbf{x})$
\end{algorithmic}
\label{alg:RM}
\end{algorithm}

\begin{figure}[H]
    \centering
    \includegraphics[width=0.5\textwidth]{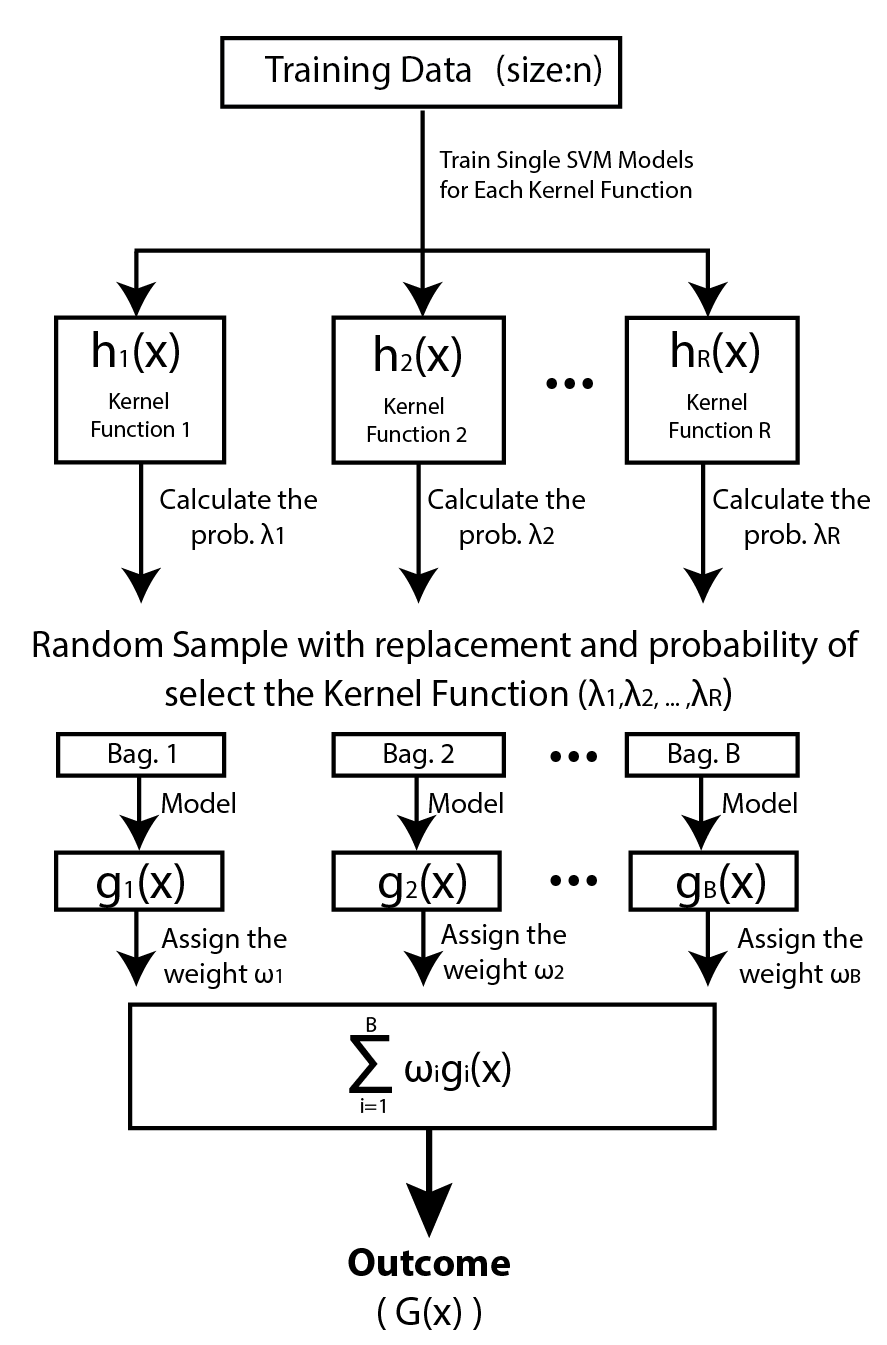}
    \caption{The graphical sketch represents all the workflow that's followed by the Random Machines.}
    \label{fig:sketch}
\end{figure}

The entire Random Machines is schematically presented in Figure \ref{fig:sketch}, where it is designed all the steps used in all cases presented in this article. %The parameter which determines the number of bootstraps samples is chosen by the user, as well as the hyperparameters from the kernel function.

\section{Artificial Data Application}

In this section simulations studies were conducted in order to evaluate the efficiency of the RM applied to binary classification tasks. The other methods compared were: \textit{linear, polynomial, Gaussian} and \textit{Laplacian} SVM, beyond their bagged versions, respectively.  A good variety between the simulated datasets is observed with three different scenarios. The dimensionality ($p$) ranges from $\{2,10,50\}$, the number of observations ($n$) ranges from $\{10,1000\}$, and the proportion's ratio between the two classes assume two values $\{0.1,0.5\}$.

The generation from the \textbf{Dataset 1} and \textbf{Dataset 2} consider continuous explanatory variables \cite{breiman1998arcing}, were the observations belonging to each class follow a multivariate distribution with their respective mean vector and covariate matrix.
The \textbf{Dataset 1} follows the configuration that instances from Class A are sampled from a normal multivariate which has mean vector $\boldsymbol{\mu}_{A}=\boldsymbol{\Vec{0}_{p}}$ and covariate matrix $\boldsymbol{\Sigma_{A}}=4\boldsymbol{I}_{p}$ and the Class B instances are sampled from a normal multivariate that has mean vector $\boldsymbol{\mu}_{B}=\boldsymbol{\Vec{4}_{p}}$ and covariate matrix $\boldsymbol{\Sigma_{A}}=\boldsymbol{I}_{p}$.
The \textbf{Dataset 2}, has the same distribution with the exception that the mean vector for Class B is given by $\boldsymbol{\mu}_{B}=\boldsymbol{\Vec{2}_{p}}$. The difference between those two datasets relies on the difficult to create the hyperplane that separate the two classes, since the \textbf{Dataset 1} has observations from each group that are further away when compared with \textbf{Dataset 2}.

The \textbf{Dataset 3} considers a classification problem in which is generated a circle uniformly distributed inside in the middle a $p$-dimensional cube. This dataset is fundamentally more complex to realize a classification, since it's has a non-linear behavior.%
The performance of each model was appraised using the following metrics. 

\textbf{Accuracy (ACC)}: it measures the ratio of correctly classified observations to total observations from the sample. It is calculated by 

\begin{equation}
    ACC=\frac{TP+TN}{TP+TN+FP+FN}
    \label{eq:acc}
\end{equation}

\textbf{Matthew's Correlation Coefficient (MCC)}: introduced by Matthews, 1975 \cite{matthews1975comparison}, is usually used to evaluate the predictions made from the model \cite{baldi2000assessing} and it is defined by,
\begin{equation}
    MCC=\frac{TP\times TN-FP\times FN}{\sqrt{(TP+FP)(TP+FN)(TN+FP)(TN+FN)}}.
    \label{eq:mcc}
\end{equation}
 It can be considered an accurate coefficient, since it penalizes the False Positive and False Negative predictions, besides being considered a better evaluator if the classes are of very different sizes \cite{boughorbel2017optimal}. It is range varies from $[-1,1]$, in which $1$ represents a perfect prediction, $0$ no better than a random choice, and $-1$ a complete reverse classification.

In order to compare directly with the accuracy, as the scales between the metrics are different, we proposed a modification to MCC. The transformation is given by $uMCC=\frac{MCC+1}{2}$ and results in a new evaluation metric: Uniform MCC (uMCC). The uMCC lies in the interval $[0,1]$, where $1$ represents a perfect prediction, $0$ no better than a random prediction.

The validation technique used was the repeated holdout with 30 repetitions with a split ratio of training-test of $70\%-30\%$. The result is summarized in Table \ref{tab:results_summary} where all possible combination of kernel functions and datasets setups are presented. It is possible to see that in most cases, the RM surpasses or equals the other methods. For instance, in Dataset 3, where the nonlinear behavior is an essential characteristic from the data, we can observe the RM overcomes the other classifiers as the dimensionality of the data increases.

% \usepackage{multirow}
% \usepackage{multirow}
% Table generated by Excel2LaTeX from sheet 'Plan1'
% Table generated by Excel2LaTeX from sheet 'Plan1'
% Table generated by Excel2LaTeX from sheet 'Plan1'

% Table generated by Excel2LaTeX from sheet 'Plan1'

% Table generated by Excel2LaTeX from sheet 'Plan1'
\begin{table}[H]
  \Huge
  \centering
  \caption{Summary of the simulation's results for the different databases.}
  \begin{adjustbox}{max width=0.78\textwidth}
    \begin{tabular}{rrrrrrrrrrrrrrrrrrrrr}
    \toprule
    \multicolumn{3}{c}{\textbf{Setup}} & \multicolumn{2}{c}{\textbf{SVM$_{lin}$}} & \multicolumn{2}{c}{\textbf{SVM$_{poly}$ }} & \multicolumn{2}{c}{\textbf{SVM$_{gaus}$}} & \multicolumn{2}{c}{\textbf{SVM$_{lap}$}} & \multicolumn{2}{c}{\textbf{BSVM$_{lin}$}} & \multicolumn{2}{c}{\textbf{BSVM$_{poly}$}} & \multicolumn{2}{c}{\textbf{BSVM$_{gau}$}} & \multicolumn{2}{c}{\textbf{BSVM$_{lap}$}} & \multicolumn{2}{c}{\textbf{RM}} \\
    \multicolumn{1}{c}{\textbf{p}} & \multicolumn{1}{c}{\textbf{Ratio}} &       & \multicolumn{1}{c}{ACC} & \multicolumn{1}{c}{uMCC} & \multicolumn{1}{c}{ACC} & \multicolumn{1}{c}{uMCC} & \multicolumn{1}{c}{ACC} & \multicolumn{1}{c}{uMCC} & \multicolumn{1}{c}{ACC} & \multicolumn{1}{c}{uMCC} & \multicolumn{1}{c}{ACC} & \multicolumn{1}{c}{uMCC} & \multicolumn{1}{c}{ACC} & \multicolumn{1}{c}{uMCC} & \multicolumn{1}{c}{ACC} & \multicolumn{1}{c}{uMCC} & \multicolumn{1}{c}{ACC} & \multicolumn{1}{c}{uMCC} & \multicolumn{1}{c}{ACC} & \multicolumn{1}{c}{uMCC} \\
    \midrule
          &       &       &       &       &       &       &       &       &       &       &       &       &       &       &       &       &       &       &       &  \\
    \multicolumn{4}{l}{\textbf{Dataset 1} (n=100)} &       &       &       &       &       &       &       &       &       &       &       &       &       &       &       &       &  \\
    \multicolumn{1}{c}{\multirow{2}[0]{*}{2}} & 0.1   &       & \multicolumn{1}{c}{1.00} & \multicolumn{1}{c}{0.98} & \multicolumn{1}{c}{1.00} & \multicolumn{1}{c}{0.98} & \multicolumn{1}{c}{0.99} & \multicolumn{1}{c}{0.96} & \multicolumn{1}{c}{0.99} & \multicolumn{1}{c}{0.97} & \multicolumn{1}{c}{1.00} & \multicolumn{1}{c}{0.98} & \multicolumn{1}{c}{1.00} & \multicolumn{1}{c}{0.98} & \multicolumn{1}{c}{0.98} & \multicolumn{1}{c}{0.92} & \multicolumn{1}{c}{0.99} & \multicolumn{1}{c}{0.95} & \multicolumn{1}{c}{1.00} & \multicolumn{1}{c}{0.97} \\
          & 0.5   &       & \multicolumn{1}{c}{0.96} & \multicolumn{1}{c}{0.96} & \multicolumn{1}{c}{0.96} & \multicolumn{1}{c}{0.96} & \multicolumn{1}{c}{0.96} & \multicolumn{1}{c}{0.96} & \multicolumn{1}{c}{0.96} & \multicolumn{1}{c}{0.96} & \multicolumn{1}{c}{0.96} & \multicolumn{1}{c}{0.96} & \multicolumn{1}{c}{0.96} & \multicolumn{1}{c}{0.96} & \multicolumn{1}{c}{0.96} & \multicolumn{1}{c}{0.96} & \multicolumn{1}{c}{0.96} & \multicolumn{1}{c}{0.96} & \multicolumn{1}{c}{0.96} & \multicolumn{1}{c}{0.96} \\
    \multicolumn{1}{c}{\multirow{2}[0]{*}{10}} & 0.1   &       & \multicolumn{1}{c}{1.00} & \multicolumn{1}{c}{0.97} & \multicolumn{1}{c}{1.00} & \multicolumn{1}{c}{0.95} & \multicolumn{1}{c}{0.91} & \multicolumn{1}{c}{0.50} & \multicolumn{1}{c}{0.91} & \multicolumn{1}{c}{0.50} & \multicolumn{1}{c}{1.00} & \multicolumn{1}{c}{0.97} & \multicolumn{1}{c}{0.99} & \multicolumn{1}{c}{0.95} & \multicolumn{1}{c}{0.91} & \multicolumn{1}{c}{0.50} & \multicolumn{1}{c}{0.91} & \multicolumn{1}{c}{0.50} & \multicolumn{1}{c}{0.99} & \multicolumn{1}{c}{0.95} \\
          & 0.5   &       & \multicolumn{1}{c}{1.00} & \multicolumn{1}{c}{1.00} & \multicolumn{1}{c}{1.00} & \multicolumn{1}{c}{1.00} & \multicolumn{1}{c}{0.95} & \multicolumn{1}{c}{0.95} & \multicolumn{1}{c}{1.00} & \multicolumn{1}{c}{1.00} & \multicolumn{1}{c}{1.00} & \multicolumn{1}{c}{1.00} & \multicolumn{1}{c}{1.00} & \multicolumn{1}{c}{1.00} & \multicolumn{1}{c}{0.95} & \multicolumn{1}{c}{0.95} & \multicolumn{1}{c}{1.00} & \multicolumn{1}{c}{1.00} & \multicolumn{1}{c}{1.00} & \multicolumn{1}{c}{1.00} \\
    \multicolumn{1}{c}{\multirow{2}[0]{*}{50}} & 0.1   &       & \multicolumn{1}{c}{1.00} & \multicolumn{1}{c}{0.97} & \multicolumn{1}{c}{1.00} & \multicolumn{1}{c}{0.97} & \multicolumn{1}{c}{0.91} & \multicolumn{1}{c}{0.50} & \multicolumn{1}{c}{0.91} & \multicolumn{1}{c}{0.50} & \multicolumn{1}{c}{1.00} & \multicolumn{1}{c}{0.97} & \multicolumn{1}{c}{1.00} & \multicolumn{1}{c}{0.97} & \multicolumn{1}{c}{0.91} & \multicolumn{1}{c}{0.50} & \multicolumn{1}{c}{0.91} & \multicolumn{1}{c}{0.50} & \multicolumn{1}{c}{1.00} & \multicolumn{1}{c}{0.97} \\
          & 0.5   &       & \multicolumn{1}{c}{1.00} & \multicolumn{1}{c}{1.00} & \multicolumn{1}{c}{1.00} & \multicolumn{1}{c}{1.00} & \multicolumn{1}{c}{0.49} & \multicolumn{1}{c}{0.56} & \multicolumn{1}{c}{1.00} & \multicolumn{1}{c}{1.00} & \multicolumn{1}{c}{1.00} & \multicolumn{1}{c}{1.00} & \multicolumn{1}{c}{1.00} & \multicolumn{1}{c}{1.00} & \multicolumn{1}{c}{0.54} & \multicolumn{1}{c}{0.54} & \multicolumn{1}{c}{1.00} & \multicolumn{1}{c}{1.00} & \multicolumn{1}{c}{1.00} & \multicolumn{1}{c}{1.00} \\
          &       &       &       &       &       &       &       &       &       &       &       &       &       &       &       &       &       &       &       &  \\
    \multicolumn{4}{l}{\textbf{Dataset 1} (n=1000)} &       &       &       &       &       &       &       &       &       &       &       &       &       &       &       &       &  \\
    \multicolumn{1}{c}{\multirow{2}[0]{*}{2}} & 0.1   &       & \multicolumn{1}{c}{0.99} & \multicolumn{1}{c}{0.96} & \multicolumn{1}{c}{0.99} & \multicolumn{1}{c}{0.97} & \multicolumn{1}{c}{0.99} & \multicolumn{1}{c}{0.97} & \multicolumn{1}{c}{0.99} & \multicolumn{1}{c}{0.97} & \multicolumn{1}{c}{0.99} & \multicolumn{1}{c}{0.96} & \multicolumn{1}{c}{0.99} & \multicolumn{1}{c}{0.97} & \multicolumn{1}{c}{0.99} & \multicolumn{1}{c}{0.97} & \multicolumn{1}{c}{0.99} & \multicolumn{1}{c}{0.97} & \multicolumn{1}{c}{0.99} & \multicolumn{1}{c}{0.97} \\
          & 0.5   &       & \multicolumn{1}{c}{0.97} & \multicolumn{1}{c}{0.97} & \multicolumn{1}{c}{0.97} & \multicolumn{1}{c}{0.97} & \multicolumn{1}{c}{0.97} & \multicolumn{1}{c}{0.97} & \multicolumn{1}{c}{0.97} & \multicolumn{1}{c}{0.97} & \multicolumn{1}{c}{0.97} & \multicolumn{1}{c}{0.97} & \multicolumn{1}{c}{0.97} & \multicolumn{1}{c}{0.97} & \multicolumn{1}{c}{0.97} & \multicolumn{1}{c}{0.97} & \multicolumn{1}{c}{0.97} & \multicolumn{1}{c}{0.97} & \multicolumn{1}{c}{0.97} & \multicolumn{1}{c}{0.97} \\
    \multicolumn{1}{c}{\multirow{2}[0]{*}{10}} & 0.1   &       & \multicolumn{1}{c}{1.00} & \multicolumn{1}{c}{1.00} & \multicolumn{1}{c}{1.00} & \multicolumn{1}{c}{1.00} & \multicolumn{1}{c}{0.90} & \multicolumn{1}{c}{0.50} & \multicolumn{1}{c}{1.00} & \multicolumn{1}{c}{1.00} & \multicolumn{1}{c}{1.00} & \multicolumn{1}{c}{1.00} & \multicolumn{1}{c}{1.00} & \multicolumn{1}{c}{1.00} & \multicolumn{1}{c}{0.90} & \multicolumn{1}{c}{0.50} & \multicolumn{1}{c}{1.00} & \multicolumn{1}{c}{1.00} & \multicolumn{1}{c}{1.00} & \multicolumn{1}{c}{1.00} \\
          & 0.5   &       & \multicolumn{1}{c}{1.00} & \multicolumn{1}{c}{1.00} & \multicolumn{1}{c}{1.00} & \multicolumn{1}{c}{1.00} & \multicolumn{1}{c}{0.99} & \multicolumn{1}{c}{0.99} & \multicolumn{1}{c}{1.00} & \multicolumn{1}{c}{1.00} & \multicolumn{1}{c}{1.00} & \multicolumn{1}{c}{1.00} & \multicolumn{1}{c}{1.00} & \multicolumn{1}{c}{1.00} & \multicolumn{1}{c}{0.98} & \multicolumn{1}{c}{0.98} & \multicolumn{1}{c}{1.00} & \multicolumn{1}{c}{1.00} & \multicolumn{1}{c}{1.00} & \multicolumn{1}{c}{1.00} \\
    \multicolumn{1}{c}{\multirow{2}[0]{*}{50}} & 0.1   &       & \multicolumn{1}{c}{1.00} & \multicolumn{1}{c}{1.00} & \multicolumn{1}{c}{1.00} & \multicolumn{1}{c}{1.00} & \multicolumn{1}{c}{0.90} & \multicolumn{1}{c}{0.50} & \multicolumn{1}{c}{0.90} & \multicolumn{1}{c}{0.50} & \multicolumn{1}{c}{1.00} & \multicolumn{1}{c}{1.00} & \multicolumn{1}{c}{1.00} & \multicolumn{1}{c}{1.00} & \multicolumn{1}{c}{0.90} & \multicolumn{1}{c}{0.50} & \multicolumn{1}{c}{0.90} & \multicolumn{1}{c}{0.50} & \multicolumn{1}{c}{1.00} & \multicolumn{1}{c}{1.00} \\
          & 0.5   &       & \multicolumn{1}{c}{1.00} & \multicolumn{1}{c}{1.00} & \multicolumn{1}{c}{1.00} & \multicolumn{1}{c}{1.00} & \multicolumn{1}{c}{0.48} & \multicolumn{1}{c}{0.51} & \multicolumn{1}{c}{1.00} & \multicolumn{1}{c}{1.00} & \multicolumn{1}{c}{1.00} & \multicolumn{1}{c}{1.00} & \multicolumn{1}{c}{1.00} & \multicolumn{1}{c}{1.00} & \multicolumn{1}{c}{0.51} & \multicolumn{1}{c}{0.51} & \multicolumn{1}{c}{1.00} & \multicolumn{1}{c}{1.00} & \multicolumn{1}{c}{1.00} & \multicolumn{1}{c}{1.00} \\
          &       &       &       &       &       &       &       &       &       &       &       &       &       &       &       &       &       &       &       &  \\
          \hline
            &       &       &       &       &       &       &       &       &       &       &       &       &       &       &       &       &       &       &       &  \\
    \multicolumn{4}{l}{\textbf{Dataset 2} (n=100)} &       &       &       &       &       &       &       &       &       &       &       &       &       &       &       &       &  \\
    \multicolumn{1}{c}{\multirow{2}[0]{*}{2}} & 0.1   &       & \multicolumn{1}{c}{0.96} & \multicolumn{1}{c}{0.86} & \multicolumn{1}{c}{0.96} & \multicolumn{1}{c}{0.86} & \multicolumn{1}{c}{0.95} & \multicolumn{1}{c}{0.80} & \multicolumn{1}{c}{0.96} & \multicolumn{1}{c}{0.86} & \multicolumn{1}{c}{0.96} & \multicolumn{1}{c}{0.88} & \multicolumn{1}{c}{0.96} & \multicolumn{1}{c}{0.87} & \multicolumn{1}{c}{0.94} & \multicolumn{1}{c}{0.76} & \multicolumn{1}{c}{0.95} & \multicolumn{1}{c}{0.82} & \multicolumn{1}{c}{0.96} & \multicolumn{1}{c}{0.87} \\
          & 0.5   &       & \multicolumn{1}{c}{0.87} & \multicolumn{1}{c}{0.88} & \multicolumn{1}{c}{0.88} & \multicolumn{1}{c}{0.88} & \multicolumn{1}{c}{0.89} & \multicolumn{1}{c}{0.89} & \multicolumn{1}{c}{0.89} & \multicolumn{1}{c}{0.89} & \multicolumn{1}{c}{0.87} & \multicolumn{1}{c}{0.88} & \multicolumn{1}{c}{0.88} & \multicolumn{1}{c}{0.88} & \multicolumn{1}{c}{0.89} & \multicolumn{1}{c}{0.89} & \multicolumn{1}{c}{0.89} & \multicolumn{1}{c}{0.89} & \multicolumn{1}{c}{0.89} & \multicolumn{1}{c}{0.89} \\
    \multicolumn{1}{c}{\multirow{2}[0]{*}{10}} & 0.1   &       & \multicolumn{1}{c}{0.99} & \multicolumn{1}{c}{0.93} & \multicolumn{1}{c}{0.95} & \multicolumn{1}{c}{0.78} & \multicolumn{1}{c}{0.91} & \multicolumn{1}{c}{0.50} & \multicolumn{1}{c}{0.91} & \multicolumn{1}{c}{0.50} & \multicolumn{1}{c}{0.98} & \multicolumn{1}{c}{0.90} & \multicolumn{1}{c}{0.96} & \multicolumn{1}{c}{0.79} & \multicolumn{1}{c}{0.91} & \multicolumn{1}{c}{0.50} & \multicolumn{1}{c}{0.91} & \multicolumn{1}{c}{0.50} & \multicolumn{1}{c}{0.96} & \multicolumn{1}{c}{0.83} \\
          & 0.5   &       & \multicolumn{1}{c}{0.93} & \multicolumn{1}{c}{0.94} & \multicolumn{1}{c}{0.94} & \multicolumn{1}{c}{0.94} & \multicolumn{1}{c}{0.85} & \multicolumn{1}{c}{0.87} & \multicolumn{1}{c}{0.98} & \multicolumn{1}{c}{0.98} & \multicolumn{1}{c}{0.94} & \multicolumn{1}{c}{0.94} & \multicolumn{1}{c}{0.94} & \multicolumn{1}{c}{0.95} & \multicolumn{1}{c}{0.81} & \multicolumn{1}{c}{0.85} & \multicolumn{1}{c}{0.98} & \multicolumn{1}{c}{0.98} & \multicolumn{1}{c}{0.96} & \multicolumn{1}{c}{0.96} \\
    \multicolumn{1}{c}{\multirow{2}[0]{*}{50}} & 0.1   &       & \multicolumn{1}{c}{1.00} & \multicolumn{1}{c}{0.97} & \multicolumn{1}{c}{0.96} & \multicolumn{1}{c}{0.81} & \multicolumn{1}{c}{0.91} & \multicolumn{1}{c}{0.50} & \multicolumn{1}{c}{0.91} & \multicolumn{1}{c}{0.50} & \multicolumn{1}{c}{1.00} & \multicolumn{1}{c}{0.97} & \multicolumn{1}{c}{0.94} & \multicolumn{1}{c}{0.69} & \multicolumn{1}{c}{0.91} & \multicolumn{1}{c}{0.50} & \multicolumn{1}{c}{0.91} & \multicolumn{1}{c}{0.50} & \multicolumn{1}{c}{0.99} & \multicolumn{1}{c}{0.92} \\
          & 0.5   &       & \multicolumn{1}{c}{1.00} & \multicolumn{1}{c}{1.00} & \multicolumn{1}{c}{0.88} & \multicolumn{1}{c}{0.90} & \multicolumn{1}{c}{0.47} & \multicolumn{1}{c}{0.54} & \multicolumn{1}{c}{0.83} & \multicolumn{1}{c}{0.88} & \multicolumn{1}{c}{1.00} & \multicolumn{1}{c}{1.00} & \multicolumn{1}{c}{0.81} & \multicolumn{1}{c}{0.84} & \multicolumn{1}{c}{0.53} & \multicolumn{1}{c}{0.53} & \multicolumn{1}{c}{0.76} & \multicolumn{1}{c}{0.77} & \multicolumn{1}{c}{1.00} & \multicolumn{1}{c}{1.00} \\
          &       &       &       &       &       &       &       &       &       &       &       &       &       &       &       &       &       &       &       &  \\
    \multicolumn{4}{l}{\textbf{Dataset 2} (n=1000)} &       &       &       &       &       &       &       &       &       &       &       &       &       &       &       &       &  \\
    \multicolumn{1}{c}{\multirow{2}[0]{*}{2}} & 0.1   &       & \multicolumn{1}{c}{0.94} & \multicolumn{1}{c}{0.79} & \multicolumn{1}{c}{0.94} & \multicolumn{1}{c}{0.82} & \multicolumn{1}{c}{0.94} & \multicolumn{1}{c}{0.81} & \multicolumn{1}{c}{0.94} & \multicolumn{1}{c}{0.81} & \multicolumn{1}{c}{0.94} & \multicolumn{1}{c}{0.79} & \multicolumn{1}{c}{0.94} & \multicolumn{1}{c}{0.82} & \multicolumn{1}{c}{0.94} & \multicolumn{1}{c}{0.81} & \multicolumn{1}{c}{0.94} & \multicolumn{1}{c}{0.82} & \multicolumn{1}{c}{0.94} & \multicolumn{1}{c}{0.82} \\
          & 0.5   &       & \multicolumn{1}{c}{0.83} & \multicolumn{1}{c}{0.83} & \multicolumn{1}{c}{0.86} & \multicolumn{1}{c}{0.86} & \multicolumn{1}{c}{0.86} & \multicolumn{1}{c}{0.87} & \multicolumn{1}{c}{0.86} & \multicolumn{1}{c}{0.86} & \multicolumn{1}{c}{0.83} & \multicolumn{1}{c}{0.83} & \multicolumn{1}{c}{0.86} & \multicolumn{1}{c}{0.86} & \multicolumn{1}{c}{0.86} & \multicolumn{1}{c}{0.86} & \multicolumn{1}{c}{0.86} & \multicolumn{1}{c}{0.86} & \multicolumn{1}{c}{0.86} & \multicolumn{1}{c}{0.86} \\
    \multicolumn{1}{c}{\multirow{2}[0]{*}{10}} & 0.1   &       & \multicolumn{1}{c}{1.00} & \multicolumn{1}{c}{0.99} & \multicolumn{1}{c}{0.99} & \multicolumn{1}{c}{0.96} & \multicolumn{1}{c}{0.90} & \multicolumn{1}{c}{0.50} & \multicolumn{1}{c}{1.00} & \multicolumn{1}{c}{0.99} & \multicolumn{1}{c}{1.00} & \multicolumn{1}{c}{0.99} & \multicolumn{1}{c}{0.99} & \multicolumn{1}{c}{0.97} & \multicolumn{1}{c}{0.90} & \multicolumn{1}{c}{0.50} & \multicolumn{1}{c}{0.98} & \multicolumn{1}{c}{0.93} & \multicolumn{1}{c}{1.00} & \multicolumn{1}{c}{0.99} \\
          & 0.5   &       & \multicolumn{1}{c}{0.98} & \multicolumn{1}{c}{0.98} & \multicolumn{1}{c}{0.99} & \multicolumn{1}{c}{0.99} & \multicolumn{1}{c}{0.92} & \multicolumn{1}{c}{0.93} & \multicolumn{1}{c}{1.00} & \multicolumn{1}{c}{1.00} & \multicolumn{1}{c}{0.98} & \multicolumn{1}{c}{0.98} & \multicolumn{1}{c}{0.99} & \multicolumn{1}{c}{0.99} & \multicolumn{1}{c}{0.91} & \multicolumn{1}{c}{0.92} & \multicolumn{1}{c}{1.00} & \multicolumn{1}{c}{1.00} & \multicolumn{1}{c}{1.00} & \multicolumn{1}{c}{1.00} \\
    \multicolumn{1}{c}{\multirow{2}[0]{*}{50}} & 0.1   &       & \multicolumn{1}{c}{1.00} & \multicolumn{1}{c}{1.00} & \multicolumn{1}{c}{1.00} & \multicolumn{1}{c}{1.00} & \multicolumn{1}{c}{0.90} & \multicolumn{1}{c}{0.50} & \multicolumn{1}{c}{0.90} & \multicolumn{1}{c}{0.50} & \multicolumn{1}{c}{1.00} & \multicolumn{1}{c}{1.00} & \multicolumn{1}{c}{1.00} & \multicolumn{1}{c}{0.99} & \multicolumn{1}{c}{0.90} & \multicolumn{1}{c}{0.50} & \multicolumn{1}{c}{0.90} & \multicolumn{1}{c}{0.50} & \multicolumn{1}{c}{1.00} & \multicolumn{1}{c}{1.00} \\
          & 0.5   &       & \multicolumn{1}{c}{1.00} & \multicolumn{1}{c}{1.00} & \multicolumn{1}{c}{1.00} & \multicolumn{1}{c}{1.00} & \multicolumn{1}{c}{0.48} & \multicolumn{1}{c}{0.50} & \multicolumn{1}{c}{1.00} & \multicolumn{1}{c}{1.00} & \multicolumn{1}{c}{1.00} & \multicolumn{1}{c}{1.00} & \multicolumn{1}{c}{1.00} & \multicolumn{1}{c}{1.00} & \multicolumn{1}{c}{0.52} & \multicolumn{1}{c}{0.52} & \multicolumn{1}{c}{1.00} & \multicolumn{1}{c}{1.00} & \multicolumn{1}{c}{1.00} & \multicolumn{1}{c}{1.00} \\
          &       &       &       &       &       &       &       &       &       &       &       &       &       &       &       &       &       &       &       &  \\ \hline
            &       &       &       &       &       &       &       &       &       &       &       &       &       &       &       &       &       &       &       &  \\
    \multicolumn{4}{l}{\textbf{Dataset 3} (n=100)} &       &       &       &       &       &       &       &       &       &       &       &       &       &       &       &       &  \\
    \multicolumn{1}{c}{\multirow{2}[0]{*}{2}} & 0.1   &       & \multicolumn{1}{c}{0.52} & \multicolumn{1}{c}{0.61} & \multicolumn{1}{c}{0.86} & \multicolumn{1}{c}{0.86} & \multicolumn{1}{c}{0.66} & \multicolumn{1}{c}{0.70} & \multicolumn{1}{c}{0.64} & \multicolumn{1}{c}{0.72} & \multicolumn{1}{c}{0.59} & \multicolumn{1}{c}{0.63} & \multicolumn{1}{c}{0.78} & \multicolumn{1}{c}{0.80} & \multicolumn{1}{c}{0.64} & \multicolumn{1}{c}{0.69} & \multicolumn{1}{c}{0.63} & \multicolumn{1}{c}{0.70} & \multicolumn{1}{c}{0.78} & \multicolumn{1}{c}{0.81} \\
          & 0.5   &       & \multicolumn{1}{c}{0.56} & \multicolumn{1}{c}{0.58} & \multicolumn{1}{c}{0.97} & \multicolumn{1}{c}{0.97} & \multicolumn{1}{c}{0.92} & \multicolumn{1}{c}{0.93} & \multicolumn{1}{c}{0.92} & \multicolumn{1}{c}{0.93} & \multicolumn{1}{c}{0.58} & \multicolumn{1}{c}{0.60} & \multicolumn{1}{c}{0.96} & \multicolumn{1}{c}{0.96} & \multicolumn{1}{c}{0.92} & \multicolumn{1}{c}{0.92} & \multicolumn{1}{c}{0.92} & \multicolumn{1}{c}{0.93} & \multicolumn{1}{c}{0.95} & \multicolumn{1}{c}{0.95} \\
    \multicolumn{1}{c}{\multirow{2}[0]{*}{10}} & 0.1   &       & \multicolumn{1}{c}{0.76} & \multicolumn{1}{c}{0.78} & \multicolumn{1}{c}{0.58} & \multicolumn{1}{c}{0.62} & \multicolumn{1}{c}{0.46} & \multicolumn{1}{c}{0.57} & \multicolumn{1}{c}{0.47} & \multicolumn{1}{c}{0.57} & \multicolumn{1}{c}{0.71} & \multicolumn{1}{c}{0.74} & \multicolumn{1}{c}{0.53} & \multicolumn{1}{c}{0.57} & \multicolumn{1}{c}{0.57} & \multicolumn{1}{c}{0.58} & \multicolumn{1}{c}{0.58} & \multicolumn{1}{c}{0.59} & \multicolumn{1}{c}{0.69} & \multicolumn{1}{c}{0.71} \\
          & 0.5   &       & \multicolumn{1}{c}{0.49} & \multicolumn{1}{c}{0.50} & \multicolumn{1}{c}{0.68} & \multicolumn{1}{c}{0.69} & \multicolumn{1}{c}{0.46} & \multicolumn{1}{c}{0.54} & \multicolumn{1}{c}{0.64} & \multicolumn{1}{c}{0.70} & \multicolumn{1}{c}{0.52} & \multicolumn{1}{c}{0.53} & \multicolumn{1}{c}{0.68} & \multicolumn{1}{c}{0.68} & \multicolumn{1}{c}{0.52} & \multicolumn{1}{c}{0.53} & \multicolumn{1}{c}{0.61} & \multicolumn{1}{c}{0.68} & \multicolumn{1}{c}{0.68} & \multicolumn{1}{c}{0.70} \\
    \multicolumn{1}{c}{\multirow{2}[0]{*}{50}} & 0.1   &       & \multicolumn{1}{c}{0.52} & \multicolumn{1}{c}{0.55} & \multicolumn{1}{c}{0.46} & \multicolumn{1}{c}{0.50} & \multicolumn{1}{c}{0.41} & \multicolumn{1}{c}{0.53} & \multicolumn{1}{c}{0.53} & \multicolumn{1}{c}{0.65} & \multicolumn{1}{c}{0.53} & \multicolumn{1}{c}{0.58} & \multicolumn{1}{c}{0.46} & \multicolumn{1}{c}{0.50} & \multicolumn{1}{c}{0.47} & \multicolumn{1}{c}{0.51} & \multicolumn{1}{c}{0.58} & \multicolumn{1}{c}{0.63} & \multicolumn{1}{c}{0.57} & \multicolumn{1}{c}{0.62} \\
          & 0.5   &       & \multicolumn{1}{c}{0.58} & \multicolumn{1}{c}{0.59} & \multicolumn{1}{c}{0.58} & \multicolumn{1}{c}{0.61} & \multicolumn{1}{c}{0.45} & \multicolumn{1}{c}{0.52} & \multicolumn{1}{c}{0.54} & \multicolumn{1}{c}{0.59} & \multicolumn{1}{c}{0.63} & \multicolumn{1}{c}{0.63} & \multicolumn{1}{c}{0.55} & \multicolumn{1}{c}{0.58} & \multicolumn{1}{c}{0.52} & \multicolumn{1}{c}{0.51} & \multicolumn{1}{c}{0.57} & \multicolumn{1}{c}{0.57} & \multicolumn{1}{c}{0.69} & \multicolumn{1}{c}{0.70} \\
          &       &       &       &       &       &       &       &       &       &       &       &       &       &       &       &       &       &       &       &  \\ 
    \multicolumn{4}{l}{\textbf{Dataset 3} (n=1000)} &       &       &       &       &       &       &       &       &       &       &       &       &       &       &       &       &  \\
    \multicolumn{1}{c}{\multirow{2}[0]{*}{2}} & 0.1   &       & \multicolumn{1}{c}{0.48} & \multicolumn{1}{c}{0.51} & \multicolumn{1}{c}{0.99} & \multicolumn{1}{c}{0.99} & \multicolumn{1}{c}{0.94} & \multicolumn{1}{c}{0.94} & \multicolumn{1}{c}{0.95} & \multicolumn{1}{c}{0.95} & \multicolumn{1}{c}{0.52} & \multicolumn{1}{c}{0.53} & \multicolumn{1}{c}{0.99} & \multicolumn{1}{c}{0.99} & \multicolumn{1}{c}{0.95} & \multicolumn{1}{c}{0.95} & \multicolumn{1}{c}{0.95} & \multicolumn{1}{c}{0.95} & \multicolumn{1}{c}{0.98} & \multicolumn{1}{c}{0.98} \\
          & 0.5   &       & \multicolumn{1}{c}{0.49} & \multicolumn{1}{c}{0.51} & \multicolumn{1}{c}{0.99} & \multicolumn{1}{c}{0.99} & \multicolumn{1}{c}{0.98} & \multicolumn{1}{c}{0.98} & \multicolumn{1}{c}{0.98} & \multicolumn{1}{c}{0.98} & \multicolumn{1}{c}{0.52} & \multicolumn{1}{c}{0.53} & \multicolumn{1}{c}{0.99} & \multicolumn{1}{c}{0.99} & \multicolumn{1}{c}{0.98} & \multicolumn{1}{c}{0.98} & \multicolumn{1}{c}{0.98} & \multicolumn{1}{c}{0.98} & \multicolumn{1}{c}{0.99} & \multicolumn{1}{c}{0.99} \\
    \multicolumn{1}{c}{\multirow{2}[0]{*}{10}} & 0.1   &       & \multicolumn{1}{c}{0.54} & \multicolumn{1}{c}{0.55} & \multicolumn{1}{c}{0.78} & \multicolumn{1}{c}{0.78} & \multicolumn{1}{c}{0.48} & \multicolumn{1}{c}{0.52} & \multicolumn{1}{c}{0.79} & \multicolumn{1}{c}{0.80} & \multicolumn{1}{c}{0.54} & \multicolumn{1}{c}{0.54} & \multicolumn{1}{c}{0.77} & \multicolumn{1}{c}{0.77} & \multicolumn{1}{c}{0.50} & \multicolumn{1}{c}{0.53} & \multicolumn{1}{c}{0.78} & \multicolumn{1}{c}{0.79} & \multicolumn{1}{c}{0.81} & \multicolumn{1}{c}{0.82} \\
          & 0.5   &       & \multicolumn{1}{c}{0.50} & \multicolumn{1}{c}{0.51} & \multicolumn{1}{c}{0.95} & \multicolumn{1}{c}{0.95} & \multicolumn{1}{c}{0.76} & \multicolumn{1}{c}{0.78} & \multicolumn{1}{c}{0.92} & \multicolumn{1}{c}{0.92} & \multicolumn{1}{c}{0.51} & \multicolumn{1}{c}{0.51} & \multicolumn{1}{c}{0.95} & \multicolumn{1}{c}{0.95} & \multicolumn{1}{c}{0.75} & \multicolumn{1}{c}{0.77} & \multicolumn{1}{c}{0.92} & \multicolumn{1}{c}{0.92} & \multicolumn{1}{c}{0.96} & \multicolumn{1}{c}{0.96} \\
    \multicolumn{1}{c}{\multirow{2}[0]{*}{50}} & 0.1   &       & \multicolumn{1}{c}{0.46} & \multicolumn{1}{c}{0.47} & \multicolumn{1}{c}{0.55} & \multicolumn{1}{c}{0.57} & \multicolumn{1}{c}{0.46} & \multicolumn{1}{c}{0.50} & \multicolumn{1}{c}{0.49} & \multicolumn{1}{c}{0.53} & \multicolumn{1}{c}{0.47} & \multicolumn{1}{c}{0.47} & \multicolumn{1}{c}{0.54} & \multicolumn{1}{c}{0.58} & \multicolumn{1}{c}{0.50} & \multicolumn{1}{c}{0.50} & \multicolumn{1}{c}{0.52} & \multicolumn{1}{c}{0.53} & \multicolumn{1}{c}{0.59} & \multicolumn{1}{c}{0.61} \\
          & 0.5   &       & \multicolumn{1}{c}{0.49} & \multicolumn{1}{c}{0.49} & \multicolumn{1}{c}{0.72} & \multicolumn{1}{c}{0.72} & \multicolumn{1}{c}{0.49} & \multicolumn{1}{c}{0.50} & \multicolumn{1}{c}{0.67} & \multicolumn{1}{c}{0.70} & \multicolumn{1}{c}{0.48} & \multicolumn{1}{c}{0.49} & \multicolumn{1}{c}{0.70} & \multicolumn{1}{c}{0.72} & \multicolumn{1}{c}{0.50} & \multicolumn{1}{c}{0.50} & \multicolumn{1}{c}{0.61} & \multicolumn{1}{c}{0.63} & \multicolumn{1}{c}{0.83} & \multicolumn{1}{c}{0.84} \\
          &       &       &       &       &       &       &       &       &       &       &       &       &       &       &       &       &       &       &       &  \\
    \bottomrule
    \end{tabular}%
    \end{adjustbox}
  \label{tab:results_summary}%
\end{table}%

\newpage

\section{Real Data Application}

Our methodology was applied on 27 real-world datasets from the UCI Repository \cite{Dua:2019} to evaluate its performance. The datasets present a wide variety in the number of observations, dimensionality, and type of data. Although, all of them represent a binary classification task. Table \ref{tab:datasets} summarizes all datasets considered. The continuous features scaled to zero mean and unit variance, in the exception of the discrete features. The validation technique is also used was the repeated holdout with 30 repetitions with a split ratio of training-test of $70\%-30\%$.

\begin{table}[H]
  \Huge
\caption{Description of the twenty seven binary data sets.
}
\begin{adjustbox}{max width=\textwidth}
\begin{tabular}{llccc|llccc}
\hline
\textbf{ID} & \multicolumn{1}{l}{\textbf{Data Set}} & \multicolumn{1}{l}{\textbf{\#Instance}} & \multicolumn{1}{l}{\textbf{\#Features}} & \multicolumn{1}{l|}{\textbf{Class Proportion}} & \textbf{ID} & \multicolumn{1}{l}{\textbf{Data Set}} & \multicolumn{1}{l}{\textbf{\#Instance}} & \multicolumn{1}{l}{\textbf{\#Feature}} & \multicolumn{1}{l}{\textbf{Class Proportion}} \\ \hline
1           & haberman                              & 306                                     & 3                                       & 81/225                                         & 15          & \textit{audit risk}                   & 775                                     & 26                                     & 305/470                                       \\
2           & heart statlog                         & 270                                     & 14                                      & 120/150                                        & 16          & \textit{adult autism}                 & 609                                     & 20                                     & 180/429                                       \\
3           & \textit{hungarian}                    & 261                                     & 10                                      & 98/163                                         & 17          & \textit{banknote}                     & 1372                                    & 4                                      & 610/762                                       \\
4           & \textit{hepatitis}                    & 80                                      & 19                                      & 33/47                                          & 18          & \textit{transfusion}                  & 748                                     & 4                                      & 178/570                                       \\
5           & \textit{liver disorders}              & 345                                     & 6                                       & 145/200                                        & 19          & \textit{caesarian}                    & 80                                      & 4                                      & 34/46                                         \\
6           & \textit{parkinsons}                   & 195                                     & 22                                      & 48/147                                         & 20          & \textit{thoraric}                     & 470                                     & 16                                     & 70/400                                        \\
7           & \textit{sonar}                        & 208                                     & 60                                      & 97/111                                         & 21          & \textit{circles}                      & 100                                     & 2                                      & 50/50                                         \\
8           & \textit{column 2C}                    & 310                                     & 6                                       & 110/210                                        & 22          & \textit{spirals}                      & 500                                     & 2                                      & 250/250                                       \\
9           & \textit{ionosphere}                   & 351                                     & 33                                      & 126/225                                        & 23          & \textit{australian}                   & 690                                     & 14                                     & 307/383                                       \\
10          & \textit{spam}                         & 4601                                    & 57                                      & 1813/2788                                      & 24          & \textit{tic tac toe}                  & 958                                     & 3                                      & 332/626                                       \\
11          & \textit{dataR2}                       & 116                                     & 9                                       & 52/64                                          & 25          & \textit{german}                       & 100                                     & 24                                     & 300/700                                       \\
12          & \textit{kidney disease}               & 155                                     & 24                                      & 41/114                                         & 26          & \textit{sick}                         & 2643                                    & 31                                     & 212/2431                                      \\
13          & \textit{clean}                        & 476                                     & 168                                     & 207/269                                        & 27          & \textit{vehicle}                      & 846                                     & 18                                     & 218/628                                       \\
14          & \textit{whosale}                      & 440                                     & 7                                       & 142/298                                        &             & \multicolumn{1}{l}{}                  & \multicolumn{1}{l}{}                    & \multicolumn{1}{l}{}                   & \multicolumn{1}{l}{}                          \\ \hline
\label{tab:datasets}
\end{tabular}
\end{adjustbox}
\end{table}

\begin{figure}[H]
    \centering
    \includegraphics[width=0.85\textwidth]{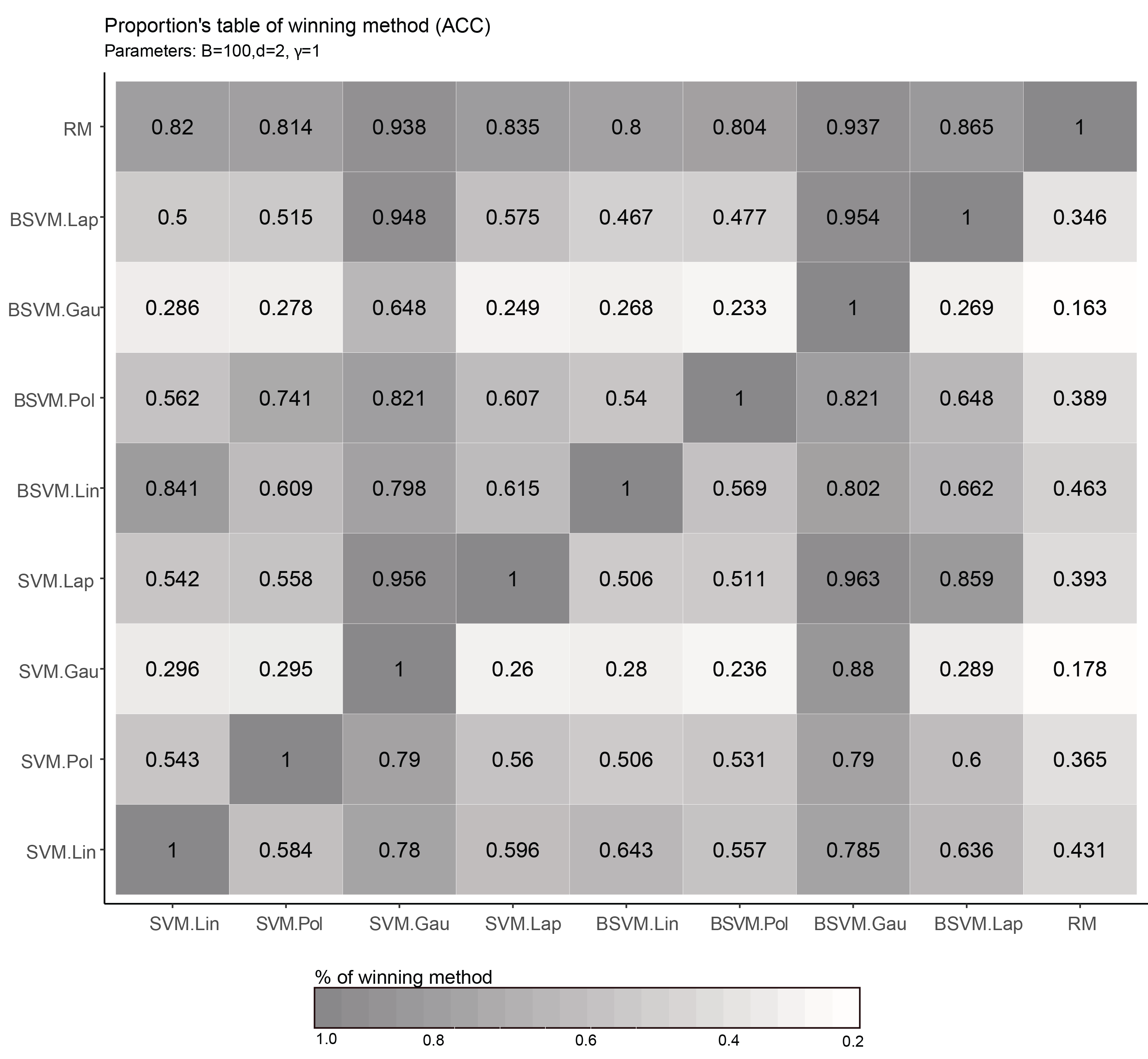}
    \caption{The chart presents the proportion of the number of times which a method have greater accuracy than the others. The proportion summarizes the applications overall 27 datasets and 30 holdout values. It is clear the superiority of the Random Machines when it is compared with the other models. }
    \label{fig:prop_table}
\end{figure}

\begin{figure}[H]
    \centering
    \includegraphics[width=0.85\textwidth]{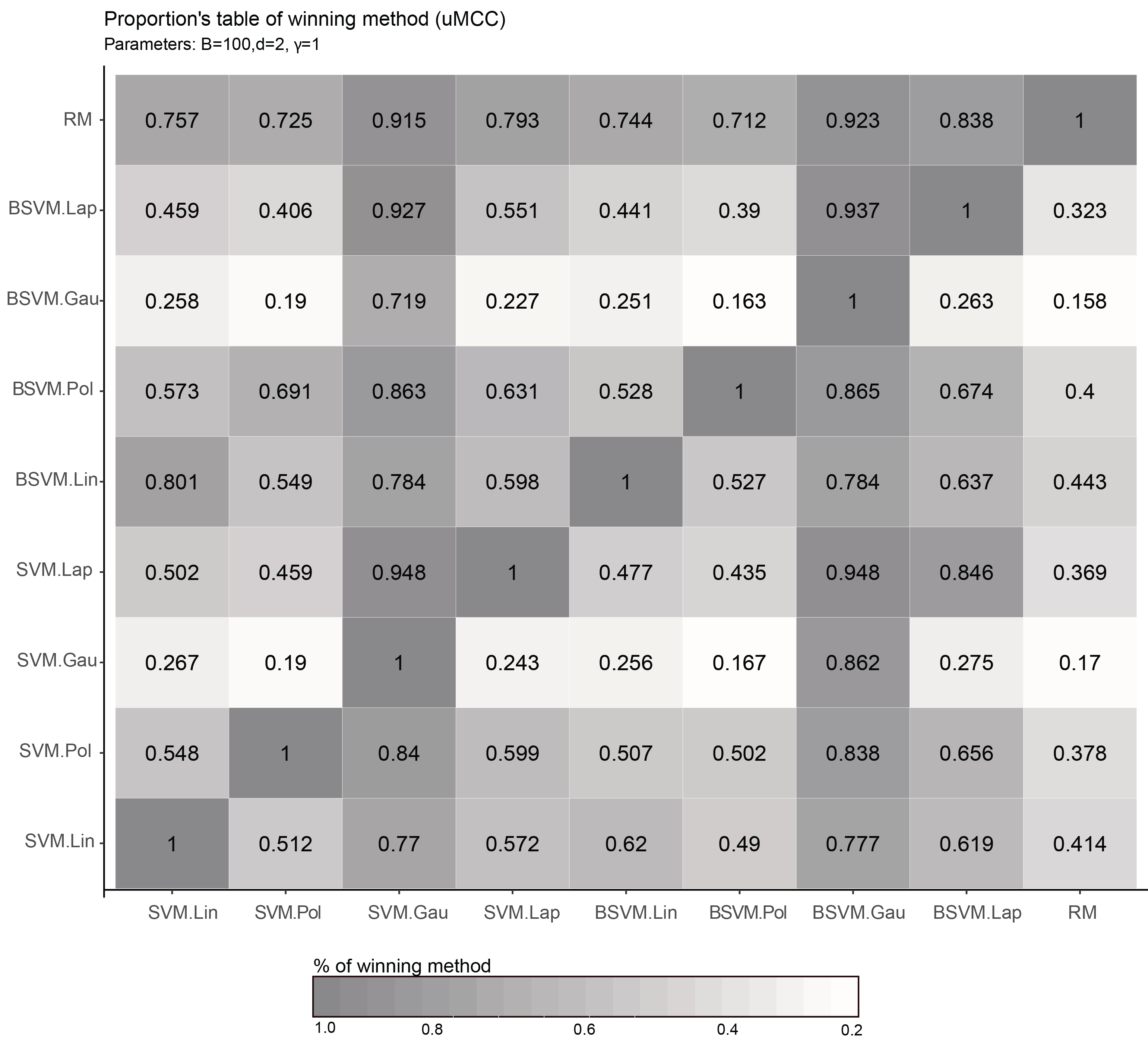}
    \caption{The chart presents the proportion of the number of times which a method have greater uMCC than the others. The proportion summarizes the applications overall 27 datasets and 30 holdout values. It is clear the superiority of the Random Machines when it is compared with the other models. }
    \label{fig:prop_table_mcc}
\end{figure}

The Random Machines was compared with the bagged SVM using each single kernel function presented in Table 1, and with the standard SVM with the same kernel functions. Without losing generality, the chosen parameters were: the cost parameter $C=1$, the number of bootstrap samples $B=100$, the degree of polynomial kernel $d=2$, and the hyperparameter $\gamma$ from the Laplacian and Gaussian kernel $\gamma=1$. The result is summarized in the Figure \ref{fig:prop_table} considering the accuracy and in the Figure \ref{fig:prop_table_mcc} considering the uMCC.

As shown in Figure \ref{fig:prop_table}, the RM demonstrates higher accuracy than the other bagged support vectors using unique kernel functions. Comparing the RM with the traditional bagged SVM, it is beaten almost 80\% of times considering the Kernel Linear Bagging, 81\% for the Kernel Polynomial Bagging, 94\% for the Gaussian Bagging, and 87\% for the Laplacian Kernel Bagging. This outcome shows off that the random weighted choice of the kernels functions improved,  generally, the accuracy of the predictions from the model. The difference is even more significant when the Random Machines is compared with the singular SVM, where the RM is more accurate $82\%$ of times considering the Kernel Linear, $81\%$ for the Kernel Polynomial, 94\% for the Gaussian Bagging, and $84\%$ for the Laplacian Kernel.

The same behavior is also observed when it is considered the Uniform Matthew's Correlation Coefficient, in which the RM present a robust superiority when compared to other methods. Analyzing the RM with the traditional bagged SVM is beaten almost $74\%$ of times considering the Kernel Linear Bagging, $71\%$ for the Kernel Polynomial Bagging, $92\%$ for the Gaussian Bagging, and $84\%$ for the Laplacian Kernel Bagging. It also happens when the RM is compared with the singular SVM, where the RM is more accurate $82\%$ of times considering the Kernel Linear, $81\%$ for the Kernel Polynomial, $94\%$ for the Gaussian Bagging, and $84\%$ for the Laplacian Kernel.

The scheme also solves the problem of the selection of best kernel function, since is not necessary to perform a grid-search among all the different functions and define which is one has lower test error, which is general framework adopted. Therefore, it is appealing that the efficiency increasing and computational cost reduction given by the technique.

\begin{figure}[H]
    \centering
    \includegraphics[width=0.85\textwidth]{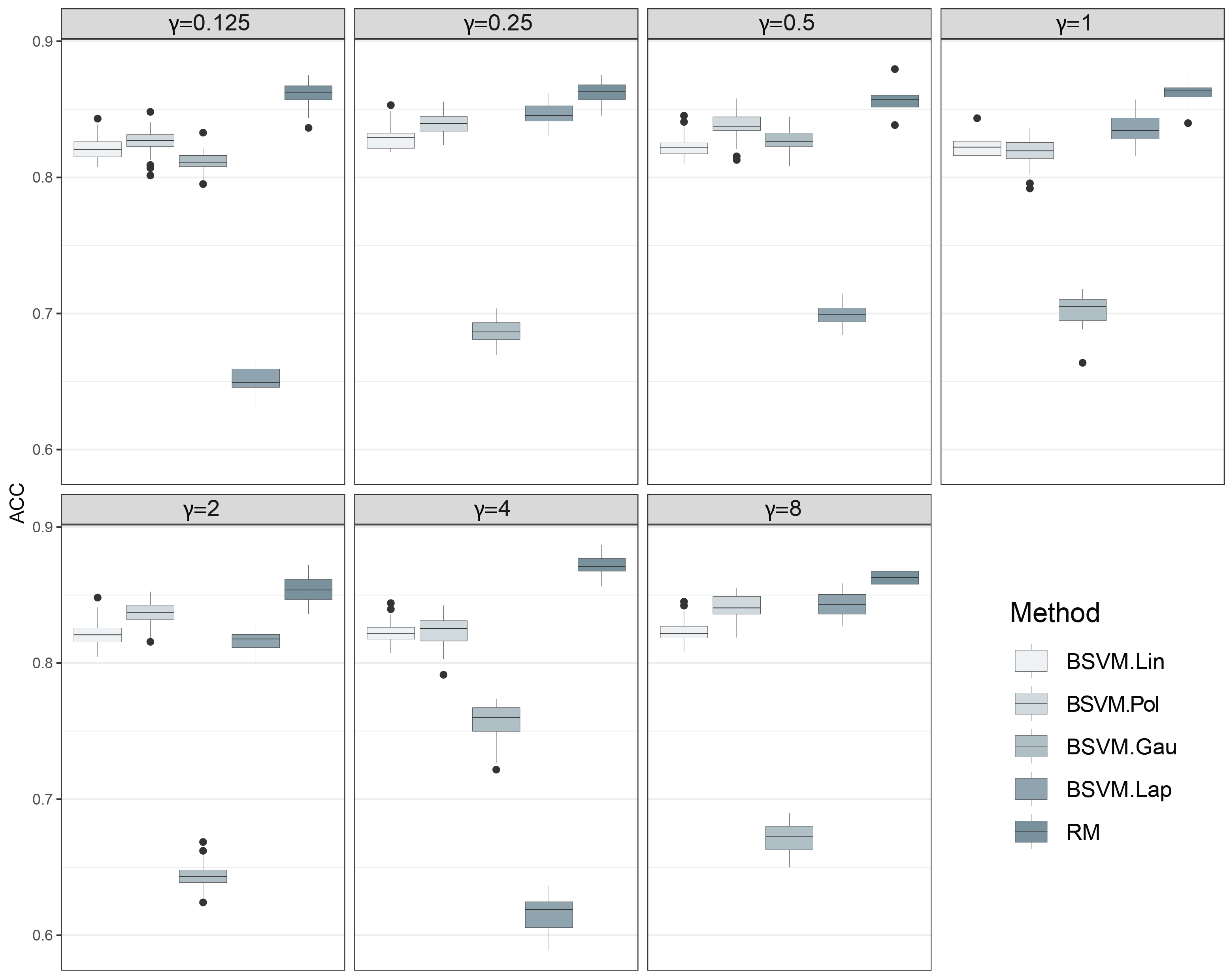}
    \caption{Summary of the ACC applied over 27 real datasets with the variation of kernel function's parameter $\gamma$.}
    \label{fig:acc_gamma}
\end{figure}

\begin{figure}[H]
    \centering
    \includegraphics[width=0.85\textwidth]{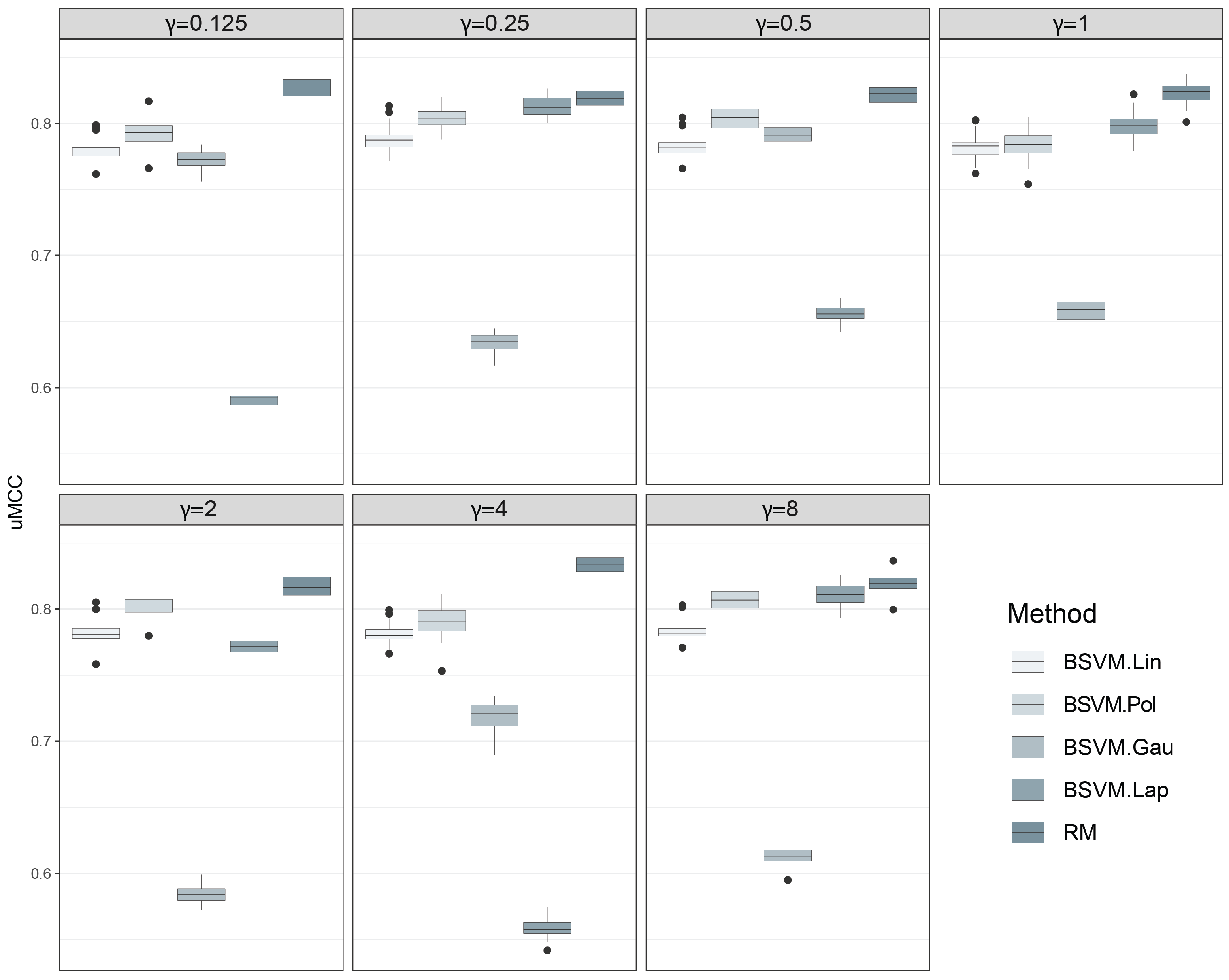}
    \caption{Summary of the uMCC applied over 27 real datasets with the variation of kernel function's parameter $\gamma$.}
    \label{fig:umcc_gamma}
\end{figure}

As the hyperparameter tuning is a remarkable question in the proceeding of the support vector machine vector, the value of $\gamma$ was changed in order to study the variation and the behavior or Random Machines when this change exists. The variation experiment relies  on the interval of values $\gamma=\{2^{-3},2^{-2},2^{-1},2^{0},2^{1},2^{2},2^{3}\}$. The result is showed in Figure \ref{fig:acc_gamma} and \ref{fig:umcc_gamma} in which it is possible to see that the RM surpassed the other bagging kernels all the times. As mentioned before the choice of these hyparameters, as the kernel function, has a direct impact on the model performance, and the results reinforce that RM gives a good and consistent result independent for all $\gamma$ values.

\section{Performance and agreement evaluation}

In this section, we justify the reason that the Random Machines is an ensemble method that can improve the predictive power for classification tasks. The main idea of the random selection of the kernel function is to select different functions that belong to a Reproducing Kernel Hilbert Space (RHKS). The reason for this choice is to aim a lower correlation between classifiers that compose the RM, and a high strength of them since these characteristics result in better results as is shown in \cite{breiman2001random}. 

The idea of the strength of a model relies on how well a model correctly predicts an new observation, while the correlation between model consists of how much they are similar. A method to estimate the correlation between classification models is to calculate the area from decision boundaries that overlaps among them \cite{turney1995bias}.Ho, \cite{timkamho1998random} defines the similarity, also called agreement, of two models as the number of observations that are equally labeled with the same class, and proposes that it can be estimated through a random sample with \textit{n} observations, by the Equation (\ref{eq:similarity}).

\begin{equation}
    \boldsymbol{\hat{s}}_{i,j}=\frac{1}{n}\sum_{k=1}^{n}f(\boldsymbol{x_{k}})
    \label{eq:similarity}
\end{equation}

where

\begin{equation*}
    f(\boldsymbol{x_{k}})=
    \begin{cases}
      1, & \text{if}\ g_{i}(\boldsymbol{x_{k}})=g_{j}(\boldsymbol{x_{k}}) \\
      0, & \text{otherwise}
    \end{cases}
  \end{equation*}

This measure called, similarity or agreement can be used as a correlation metric between models.

In order to evaluate the correlation and strength of the RM in comparison with the traditional bagged version of SVM, the method was applied over the \textit{Circles} database that was generated under the same configuration of \textbf{Dataset 3} presented in Section 5. The similarity of each method was estimated using the average of the similarity $\boldsymbol{\hat{s}_{i,j}}$, $\forall i \neq j$ and $i,j=1,\dots,B$, over fixed \textit{k} points generated by a Monte Carlo's simulation.  The accuracy was used in order to measure the strength of the model.

The dataset was modified in three configurations, changing the dimension \textit{p} in a range corresponding to \textit{p}=\{2, 10, 30, 50\}. The average similarity, that can also be called as \textit{agreeement} of the model \cite{timkamho1998random}, was calculated using \textit{k} observations, where \textit{k}=1000 $\times$ \textit{p}. Both accuracy and agreement were calculated using a 30 Repeated Holdout validation set with split ratio of 70-30\% training-test. The parameters of the methods were: B=100, $\gamma=1$, $C=1$.

\begin{figure}[H]
    \centering
    \includegraphics[width=0.75\textwidth]{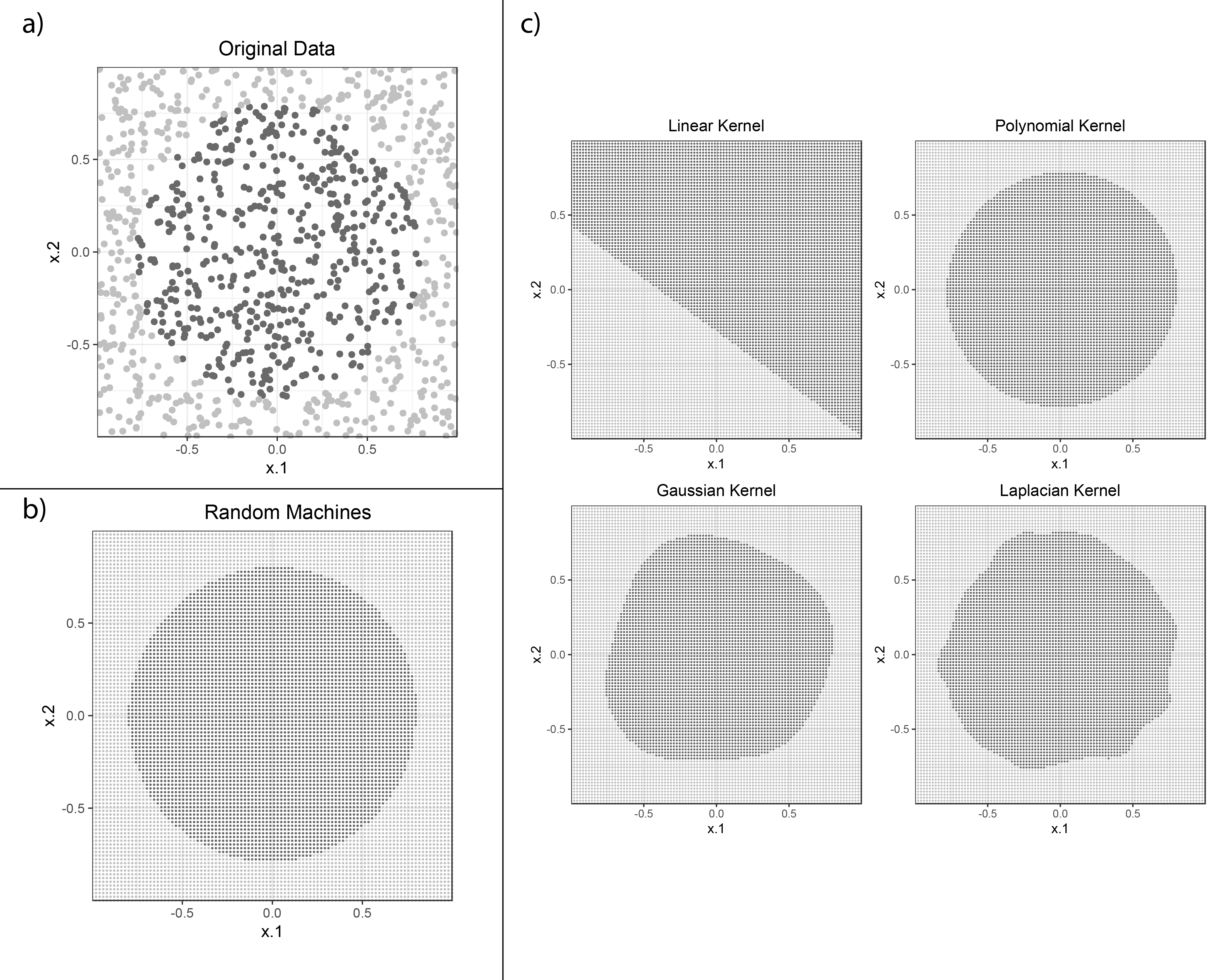}
    \caption{The figure shows the \textit{circles} database where \textit{p=2}. The panel (a) show all the observations with the class associated with each color. The panel (b) present the decision region given by the RM. The panel (c) reveals the diversity of decision regions produced by each kernel function of bootstrap models that composes the RM.}
    \label{fig:kernel_grid}
\end{figure}

One of the main results can be represented in the Figure \ref{fig:kernel_grid} where the \textit{circles} database with \textit{p}=2 was used as example. In the Figure \ref{fig:kernel_grid} (a) the plot of observations, which in each color represents a class. The panel (b) represents the final decision boundary of the RM, showing that the model captures the behavior from the observations. The panel (c) shows examples of the decision region generated by a bootstrap model $g_i$ for each kernel. 

It is clear that different kernel functions used in each SVM model produce diverse decision boundaries, and that difference implies in a reduction of the correlation, resulting at the decreasing of generalization error.

All the results are summarized in Table \ref{tab:acc_agr} where it is presented the mean accuracy and agreement for each dataset for all configurations of the \textit{circles}.

% Table generated by Excel2LaTeX from sheet 'Plan1'
\begin{table}[htbp]
  \centering
  \caption{Summary of Accuracy and Agreement measure to each method}
    \begin{adjustbox}{max width=0.8\textwidth}

    \begin{tabular}{c|cccccccccc}
    \multicolumn{1}{c|}{\multirow{2}[3]{*}{\textbf{Circles Dataset
    }}} & \multicolumn{10}{c}{\textbf{Method}} \\
\cmidrule{2-11}          & \multicolumn{2}{c}{\textbf{BSVM.Lin}} & \multicolumn{2}{c}{\textbf{BSVM.Pol}} & \multicolumn{2}{c}{\textbf{BSVM.Gau}} & \multicolumn{2}{c}{\textbf{BSVM.Lap}} & \multicolumn{2}{c}{\textbf{RM}} \\
    \midrule
    \textbf{p} & \textbf{ACC} & \multicolumn{1}{c}{\textbf{AGR}} & \textbf{ACC} & \multicolumn{1}{c}{\textbf{AGR}} & \textbf{ACC} & \multicolumn{1}{c}{\textbf{AGR}} & \textbf{ACC} & \multicolumn{1}{c}{\textbf{AGR}} & \textbf{ACC} & \textbf{AGR} \\
    \midrule
    2     & 0.54  & 0.59  & 0.98  & 0.98  & 0.97  & 0.97  & 0.97  & 0.97  & 0.99  & 0.96 \\
    10    & 0.49  & 0.64  & 0.95  & 0.92  & 0.74  & 0.72  & 0.91  & 0.91  & 0.96  & 0.84 \\
    30    & 0.49  & 0.49  & 0.78  & 0.78  & 0.51  & 0.59  & 0.87  & 0.87  & 0.94  & 0.67 \\
    50    & 0.55  & 0.67  & 0.71  & 0.71  & 0.49  & 0.61  & 0.57  & 0.62  & 0.79  & 0.62 \\
    \bottomrule
    \end{tabular}%
    \end{adjustbox}
  \label{tab:acc_agr}%
\end{table}%

In general, it is remarkable that the higher predictive capacity of the RM when compared to the other methods in all cases. Moreover, beyond the great accuracy, it is possible to see that the RM shows simultaneously a lower agreement when compared with the other methods that have an excellent performance. Although sometimes the BSVM.Lin and BSVM.Gau produce a desirable low agreement, they are considered weaker, since they have a lower accuracy when compared with the others.
As \cite{timkamho1998random} discuss, the accuracy of the models affects the agreement and vice-versa, and optimize both values simultaneously can be a challenging task. Generally, models with high accuracy also result in large agreement values, as we can see in the results exhibited in Table \ref{tab:acc_agr}. On another hand, small values of accuracy produce lower agreement measures among models. However, it is clear to notice that the Random Machines it is capable create a better classifier (low generalization error) with both characteristics: low correlation and reliable strength.

\begin{figure}[H]
    \centering
    \includegraphics[width=0.9\textwidth]{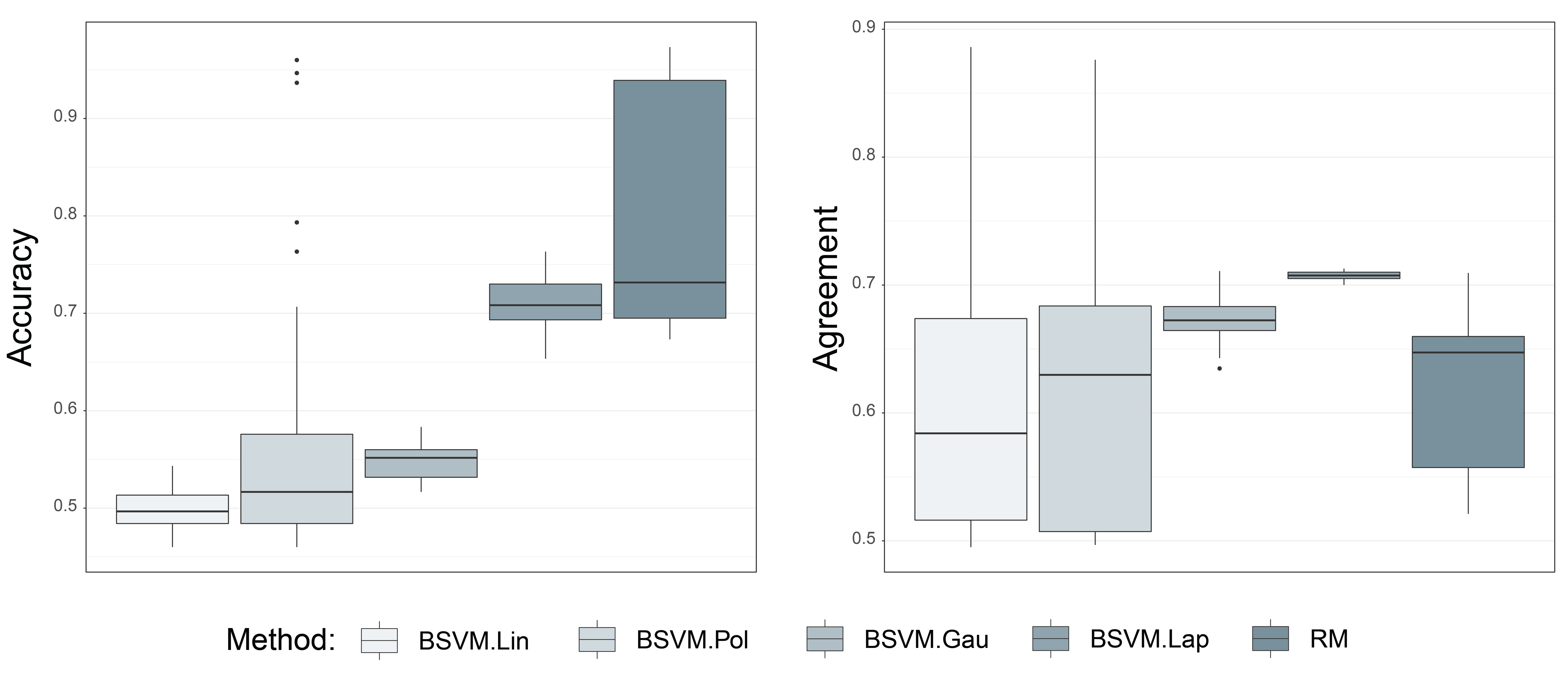}
    \caption{Boxplots of the accuracy and agreement for each method.}
    \label{fig:acc_and_agr}
\end{figure}

These proprieties become better with higher dimensions as they can be observed in Table \ref{tab:acc_agr}. This difference is showed in Figure \ref{fig:acc_and_agr} that display the boxplots of the accuracy and agreement for each method, reinforcing even more than the RM has both better proprieties than the other ones.

\section{Final Considerations}

{
The main contribution of this paper is to propose a novel learning method to do ensemble using Support Vector Machine models that can enhance the accuracy from the conventional BSVM, and solve the problem choosing the best kernel function that should be used. Through the Random Machines, the combination of different SVM using the different kernel functions states an approach that avoids the expensive computational cost of doing a grid search between the kernel functions, besides improve the accuracy. Furthermore, our results show a good behavior with different kernel hyperparameters in RM, that provides a bagged-weighted support vector model with free kernel choice. In this way, as SVM is one of the most important and an essential method in machine learning with high-performance capacity and power of generalization, the RM method can be viewed as an extension of traditional SVM, giving an alternative solution to the hyperparameters choice problem.  This methodology can be explored in many other contexts, as well as be applied to any practical machine learning problem. Future theoretical studies may be done regarding computational cost, comparison with other traditional machine learning methods, and the use of other and more kernel functions as well as other weights in the bagging phase.
}

%\appendix

%\label{appendix-sec1}

%% Sample text. Sample text. Sample text. Sample text. Sample text. Sample text. 
%% Sample text. Sample text. Sample text. Sample text. Sample text. Sample text. 
%% Sample text. 

%% References


\bibliography{mybibfile}

\begin{thebibliography}{10}
\expandafter\ifx\csname url\endcsname\relax
  \def\url#1{\texttt{#1}}\fi
\expandafter\ifx\csname urlprefix\endcsname\relax\def\urlprefix{URL }\fi
\expandafter\ifx\csname href\endcsname\relax
  \def\href#1#2{#2} \def\path#1{#1}\fi

\bibitem{sato2019machine}
M.~Sato, K.~Morimoto, S.~Kajihara, R.~Tateishi, S.~Shiina, K.~Koike, Y.~Yatomi,
  Machine-learning approach for the development of a novel predictive model for
  the diagnosis of hepatocellular carcinoma., Scientific reports 9~(1) (2019)
  7704--7704.

\bibitem{mokgonyane2019automatic}
T.~B. Mokgonyane, T.~J. Sefara, T.~I. Modipa, M.~M. Mogale, M.~J. Manamela,
  P.~J. Manamela, Automatic speaker recognition system based on machine
  learning algorithms, in: 2019 Southern African Universities Power Engineering
  Conference/Robotics and Mechatronics/Pattern Recognition Association of South
  Africa (SAUPEC/RobMech/PRASA), IEEE, 2019, pp. 141--146.

\bibitem{burdisso2019text}
S.~G. Burdisso, M.~Errecalde, M.~Montes-y G{\'o}mez, A text classification
  framework for simple and effective early depression detection over social
  media streams, Expert Systems with Applications.

\bibitem{kim2005dimension}
H.~Kim, P.~Howland, H.~Park, Dimension reduction in text classification with
  support vector machines, Journal of Machine Learning Research 6~(Jan) (2005)
  37--53.

\bibitem{dighe2018detection}
D.~Dighe, S.~Patil, S.~Kokate, Detection of credit card fraud transactions
  using machine learning algorithms and neural networks: A comparative study,
  in: 2018 Fourth International Conference on Computing Communication Control
  and Automation (ICCUBEA), IEEE, 2018, pp. 1--6.

\bibitem{cortes1995support}
C.~Cortes, V.~Vapnik, Support-vector networks, Machine learning 20~(3) (1995)
  273--297.

\bibitem{smola2000advances}
A.~J. Smola, P.~J. Bartlett, D.~Schuurmans, B.~Sch{\"o}lkopf, M.~I. Jordan,
  et~al., Advances in large margin classifiers, MIT press, 2000.

\bibitem{cueto2019comparative}
N.~Cueto-L{\'o}pez, M.~T. Garc{\'\i}a-Ord{\'a}s, V.~D{\'a}vila-Batista,
  V.~Moreno, N.~Aragon{\'e}s, R.~Alaiz-Rodr{\'\i}guez, A comparative study on
  feature selection for a risk prediction model for colorectal cancer, Computer
  Methods and Programs in Biomedicine.

\bibitem{thanh2018comparison}
P.~Thanh~Noi, M.~Kappas, Comparison of random forest, k-nearest neighbor, and
  support vector machine classifiers for land cover classification using
  sentinel-2 imagery, Sensors 18~(1) (2018) 18.

\bibitem{shah2018performance}
S.~A.~R. Shah, B.~Issac, Performance comparison of intrusion detection systems
  and application of machine learning to snort system, Future Generation
  Computer Systems 80 (2018) 157--170.

\bibitem{van2007improved}
M.~Van~Wezel, R.~Potharst, Improved customer choice predictions using ensemble
  methods, European Journal of Operational Research 181~(1) (2007) 436--452.

\bibitem{breiman1996bagging}
L.~Breiman, Bagging predictors, Machine learning 24~(2) (1996) 123--140.

\bibitem{freund1999short}
Y.~Freund, R.~Schapire, N.~Abe, A short introduction to boosting,
  Journal-Japanese Society For Artificial Intelligence 14~(771-780) (1999)
  1612.

\bibitem{kim2002support}
H.-C. Kim, S.~Pang, H.-M. Je, D.~Kim, S.-Y. Bang, Support vector machine
  ensemble with bagging, in: International Workshop on Support Vector Machines,
  Springer, 2002, pp. 397--408.

\bibitem{wang2009empirical}
S.-j. Wang, A.~Mathew, Y.~Chen, L.-f. Xi, L.~Ma, J.~Lee, Empirical analysis of
  support vector machine ensemble classifiers, Expert Systems with applications
  36~(3) (2009) 6466--6476.

\bibitem{huang2017svm}
M.-W. Huang, C.-W. Chen, W.-C. Lin, S.-W. Ke, C.-F. Tsai, Svm and svm ensembles
  in breast cancer prediction, PloS one 12~(1) (2017) e0161501.

\bibitem{wang2018support}
H.~Wang, B.~Zheng, S.~W. Yoon, H.~S. Ko, A support vector machine-based
  ensemble algorithm for breast cancer diagnosis, European Journal of
  Operational Research 267~(2) (2018) 687--699.

\bibitem{zhou2010least}
L.~Zhou, K.~K. Lai, L.~Yu, Least squares support vector machines ensemble
  models for credit scoring, Expert Systems with Applications 37~(1) (2010)
  127--133.

\bibitem{tong2013ensemble}
M.~Tong, K.-H. Liu, C.~Xu, W.~Ju, An ensemble of svm classifiers based on gene
  pairs, Computers in biology and medicine 43~(6) (2013) 729--737.

\bibitem{pham2018bagging}
B.~T. Pham, D.~T. Bui, I.~Prakash, Bagging based support vector machines for
  spatial prediction of landslides, Environmental earth sciences 77~(4) (2018)
  146.

\bibitem{gordon2005improved}
J.~J. Gordon, M.~W. Towsey, J.~M. Hogan, S.~A. Mathews, P.~Timms, Improved
  prediction of bacterial transcription start sites, Bioinformatics 22~(2)
  (2005) 142--148.

\bibitem{lei2006ensemble}
Z.~Lei, Y.~Yang, Z.~Wu, Ensemble of support vector machine for text-independent
  speaker recognition, Int. J. Comput. Sci. Networks Secur 6~(5) (2006)
  163--167.

\bibitem{pang2003membership}
S.~Pang, D.~Kim, S.~Y. Bang, Membership authentication in the dynamic group by
  face classification using svm ensemble, Pattern Recognition Letters 24~(1-3)
  (2003) 215--225.

\bibitem{jebara2004multi}
T.~Jebara, Multi-task feature and kernel selection for svms, in: Proceedings of
  the twenty-first international conference on Machine learning, ACM, 2004,
  p.~55.

\bibitem{bergstra2012random}
J.~Bergstra, Y.~Bengio, Random search for hyper-parameter optimization, Journal
  of Machine Learning Research 13~(Feb) (2012) 281--305.

\bibitem{bergstra2011algorithms}
J.~S. Bergstra, R.~Bardenet, Y.~Bengio, B.~K{\'e}gl, Algorithms for
  hyper-parameter optimization, in: Advances in neural information processing
  systems, 2011, pp. 2546--2554.

\bibitem{kirkpatrick1983optimization}
S.~Kirkpatrick, C.~D. Gelatt, M.~P. Vecchi, Optimization by simulated
  annealing, science 220~(4598) (1983) 671--680.

\bibitem{williams2001using}
C.~K. Williams, M.~Seeger, Using the nystr{\"o}m method to speed up kernel
  machines, in: Advances in neural information processing systems, 2001, pp.
  682--688.

\bibitem{smola2000sparse}
A.~J. Smola, B.~Sch{\"o}lkopf, Sparse greedy matrix approximation for machine
  learning.

\bibitem{sun2018but}
Y.~Sun, A.~Gilbert, A.~Tewari, But how does it work in theory? linear svm with
  random features, in: Advances in Neural Information Processing Systems, 2018,
  pp. 3379--3388.

\bibitem{li2016fast}
Z.~Li, T.~Yang, L.~Zhang, R.~Jin, Fast and accurate refined nystr{\"o}m-based
  kernel svm, in: Thirtieth AAAI Conference on Artificial Intelligence, 2016.

\bibitem{boser1992training}
B.~E. Boser, I.~M. Guyon, V.~N. Vapnik, A training algorithm for optimal margin
  classifiers, in: Proceedings of the fifth annual workshop on Computational
  learning theory, ACM, 1992, pp. 144--152.

\bibitem{fletcher1987practical}
R.~Fletcher, Practical methods of optimization john wiley \& sons, New York 80.

\bibitem{courant1953methods}
R.~Courant, D.~Hilbert, Methods of mathematical physics, vol. i, interscience
  publ, Inc., New York (1953) 106.

\bibitem{hussain2011comparison}
M.~Hussain, S.~K. Wajid, A.~Elzaart, M.~Berbar, A comparison of svm kernel
  functions for breast cancer detection, in: 2011 Eighth International
  Conference Computer Graphics, Imaging and Visualization, IEEE, 2011, pp.
  145--150.

\bibitem{min2005bankruptcy}
J.~H. Min, Y.-C. Lee, Bankruptcy prediction using support vector machine with
  optimal choice of kernel function parameters, Expert systems with
  applications 28~(4) (2005) 603--614.

\bibitem{chapelle2000model}
O.~Chapelle, V.~Vapnik, Model selection for support vector machines, in:
  Advances in neural information processing systems, 2000, pp. 230--236.

\bibitem{ayat2005automatic}
N.-E. Ayat, M.~Cheriet, C.~Y. Suen, Automatic model selection for the
  optimization of svm kernels, Pattern Recognition 38~(10) (2005) 1733--1745.

\bibitem{wu2009novel}
C.-H. Wu, G.-H. Tzeng, R.-H. Lin, A novel hybrid genetic algorithm for kernel
  function and parameter optimization in support vector regression, Expert
  Systems with Applications 36~(3) (2009) 4725--4735.

\bibitem{friedrichs2005evolutionary}
F.~Friedrichs, C.~Igel, Evolutionary tuning of multiple svm parameters,
  Neurocomputing 64 (2005) 107--117.

\bibitem{cherkassky2004practical}
V.~Cherkassky, Y.~Ma, Practical selection of svm parameters and noise
  estimation for svm regression, Neural networks 17~(1) (2004) 113--126.

\bibitem{lin2008support}
H.-T. Lin, L.~Li, Support vector machinery for infinite ensemble learning,
  Journal of Machine Learning Research 9~(Feb) (2008) 285--312.

\bibitem{claesen2014ensemblesvm}
M.~Claesen, F.~De~Smet, J.~A. Suykens, B.~De~Moor, Ensemblesvm: a library for
  ensemble learning using support vector machines, The Journal of Machine
  Learning Research 15~(1) (2014) 141--145.

\bibitem{breiman1998arcing}
L.~Breiman, et~al., Arcing classifier (with discussion and a rejoinder by the
  author), The annals of statistics 26~(3) (1998) 801--849.

\bibitem{matthews1975comparison}
B.~W. Matthews, Comparison of the predicted and observed secondary structure of
  t4 phage lysozyme, Biochimica et Biophysica Acta (BBA)-Protein Structure
  405~(2) (1975) 442--451.

\bibitem{baldi2000assessing}
P.~Baldi, S.~Brunak, Y.~Chauvin, C.~A. Andersen, H.~Nielsen, Assessing the
  accuracy of prediction algorithms for classification: an overview,
  Bioinformatics 16~(5) (2000) 412--424.

\bibitem{boughorbel2017optimal}
S.~Boughorbel, F.~Jarray, M.~El-Anbari, Optimal classifier for imbalanced data
  using matthews correlation coefficient metric, PloS one 12~(6) (2017)
  e0177678.

\bibitem{Dua:2019}
D.~Dua, C.~Graff, \href{http://archive.ics.uci.edu/ml}{{UCI} machine learning
  repository} (2017).
\newline\urlprefix\url{http://archive.ics.uci.edu/ml}

\bibitem{breiman2001random}
L.~Breiman, Random forests, Machine learning 45~(1) (2001) 5--32.

\bibitem{turney1995bias}
P.~Turney, Bias and the quantification of stability, Machine Learning 20~(1-2)
  (1995) 23--33.

\bibitem{timkamho1998random}
T.~K. Ho, The random subspace method for constructing decision forests, IEEE
  Trans. Pattern Anal. Mach. Intell 20~(8) (1998) 1--22.

\end{thebibliography}

\end{document}